\documentclass[fleqn,10pt]{wlscirep}
\usepackage[utf8]{inputenc}
\usepackage[T1]{fontenc}

\usepackage{geometry}
\usepackage{graphicx}
\usepackage{amssymb}
\usepackage{caption}
\usepackage{subcaption}
\usepackage{enumitem}
\usepackage{mathtools}
\usepackage[all]{xy}
\usepackage{xcolor}
\usepackage{soul}
\usepackage{bm}
\usepackage{microtype}
\usepackage{tikz}
\usetikzlibrary{positioning, fit, arrows.meta, shapes}

\usepackage{bbold}

\usepackage{MnSymbol}

\graphicspath{{figures/}}

\numberwithin{equation}{section}

\newtheorem{theorem}{Theorem}[section]
\newtheorem{proposition}{Proposition}[section]
\newtheorem{remark}{Remark}[section]

\newtheorem{definition}{Definition}[section]

\newtheorem{lemma}{Lemma}[section]
\newtheorem{example}{Example}[section]

\usepackage{dutchcal}

\newlength{\tfwidth}
\newlength{\tfheight}
\newlength{\tfxa}
\newlength{\tfxb}
\newlength{\tfya}
\newlength{\tfyb}
\newcommand{\trimFigWithBox}[6]{%
\setlength\fboxsep{0pt}%
\setlength\fboxrule{1.0pt}
\fbox{\includegraphics[width=#2, clip, trim=#3 #4 #5 #6]{#1}}%
}
\newcommand{\trimFigNoBox}[6]{%
\setlength\fboxsep{1pt}
\setlength\fboxrule{0.0pt}
\fbox{\includegraphics[width=#2, clip, trim=#3 #4 #5 #6]{#1}}%
}
\newcommand{\trimFigHeightWithBox}[6]{%
\setlength\fboxsep{0pt}%
\setlength\fboxrule{1.0pt}
\fbox{\includegraphics[height=#2, clip, trim=#3 #4 #5 #6]{#1}}%
}
\newcommand{\trimFigHeightNoBox}[6]{%
\setlength\fboxsep{1pt}
\setlength\fboxrule{0.0pt}
\fbox{\includegraphics[height=#2, clip, trim=#3 #4 #5 #6]{#1}}%
}

\newsavebox\figBox

\newcommand{\trimw}[6]{%
\sbox\figBox{\includegraphics{#1}}
\setlength{\tfwidth}{\the\wd\figBox}
\setlength{\tfheight}{\the\ht\figBox}
\setlength{\tfxa}{\tfwidth*\real{#3}}%
\setlength{\tfxb}{\tfwidth*\real{#4}}%
\setlength{\tfya}{\tfheight*\real{#5}}%
\setlength{\tfyb}{\tfheight*\real{#6}}%
\trimFigNoBox{#1}{#2}{\tfxa}{\tfya}{\tfxb}{\tfyb}%
}

\newcommand{\trimwb}[6]{%

\sbox\figBox{\includegraphics{#1}}
\setlength{\tfwidth}{\the\wd\figBox}
\setlength{\tfheight}{\the\ht\figBox}
\setlength{\tfxa}{\tfwidth*\real{#3}}%
\setlength{\tfxb}{\tfwidth*\real{#4}}%
\setlength{\tfya}{\tfheight*\real{#5}}%
\setlength{\tfyb}{\tfheight*\real{#6}}%
\trimFigWithBox{#1}{#2}{\tfxa}{\tfya}{\tfxb}{\tfyb}%
}

\newcommand{\trimh}[6]{%
\sbox\figBox{\includegraphics{#1}}
\setlength{\tfwidth}{\the\wd\figBox}
\setlength{\tfheight}{\the\ht\figBox}
\setlength{\tfxa}{\tfwidth*\real{#3}}%
\setlength{\tfxb}{\tfwidth*\real{#4}}%
\setlength{\tfya}{\tfheight*\real{#5}}%
\setlength{\tfyb}{\tfheight*\real{#6}}%
\trimFigHeightNoBox{#1}{#2}{\tfxa}{\tfya}{\tfxb}{\tfyb}%
}

\newcommand{\trimhb}[6]{%

\sbox\figBox{\includegraphics{#1}}
\setlength{\tfwidth}{\the\wd\figBox}
\setlength{\tfheight}{\the\ht\figBox}
\setlength{\tfxa}{\tfwidth*\real{#3}}%
\setlength{\tfxb}{\tfwidth*\real{#4}}%
\setlength{\tfya}{\tfheight*\real{#5}}%
\setlength{\tfyb}{\tfheight*\real{#6}}%
\trimFigHeightWithBox{#1}{#2}{\tfxa}{\tfya}{\tfxb}{\tfyb}%
}
\title{Approximation of Nearly-Periodic Symplectic Maps \\ via Structure-Preserving Neural Networks}

\author[1,*]{Valentin Duruisseaux}
\author[2]{Joshua W. Burby}
\author[2]{Qi Tang}
\affil[1]{Department of Mathematics, University of California San Diego,  La Jolla, CA 92093}
\affil[2]{Theoretical Division, Los Alamos National Laboratory, Los Alamos, NM 87545}

\affil[*]{Corresponding author: vduruiss@ucsd.edu}

\begin{abstract}
	A continuous-time dynamical system with parameter $\varepsilon$ is nearly-periodic if all its trajectories are periodic with nowhere-vanishing angular frequency as $\varepsilon$ approaches 0. Nearly-periodic maps are discrete-time analogues of nearly-periodic systems, defined as parameter-dependent diffeomorphisms that limit to rotations along a circle action, and they admit formal $U(1)$ symmetries to all orders when the limiting rotation is non-resonant. For Hamiltonian nearly-periodic maps on exact presymplectic manifolds, the formal $U(1)$ symmetry gives rise to a discrete-time adiabatic invariant. In this paper, we construct a novel structure-preserving neural network to approximate nearly-periodic symplectic maps. This neural network architecture, which we call symplectic gyroceptron, ensures that the resulting surrogate map is nearly-periodic and symplectic, and that it gives rise to a discrete-time adiabatic invariant and a long-time stability. This new structure-preserving neural network provides a promising architecture for surrogate modeling of non-dissipative dynamical systems that automatically steps over short timescales without introducing spurious instabilities. 
\end{abstract}

\begin{document}

\flushbottom
\maketitle
%
%
\thispagestyle{empty}

\section{Introduction} 

Dynamical systems evolve according to the laws of physics, which can usually be described using differential equations. By solving these differential equations, it is possible to predict the future states of the dynamical system. Identifying accurate and efficient dynamic models based on observed trajectories is thus critical for the analysis, simulation and control of dynamical systems. We consider here the problem of learning dynamics: given a dataset of trajectories followed by a dynamical system, we wish to infer the dynamical law responsible for these trajectories and then possibly use that law to predict the evolution of similar systems in different initial states. We are particularly interested in the surrogate modeling problem: the underlying dynamical system is known, but traditional simulations are either too slow or expensive for some optimization task. This problem can be addressed by learning a less expensive, but less accurate surrogate for the simulations.

Models obtained from first principles are extensively used across science and engineering. Unfortunately, due to incomplete knowledge, these models based on physical laws tend to over-simplify or incorrectly describe the underlying structure of the dynamical systems, and usually lead to high bias and modeling errors that cannot be corrected by optimizing over the few parameters in the models. 

Deep learning architectures can provide very expressive models for function approximation, and have proven very effective in numerous contexts~\cite{karniadakis2021physics, BurbyHenon, Jin2020}. Unfortunately, standard non-structure-preserving neural networks struggle to learn the symmetries and conservation laws underlying dynamical systems, and as a result do not generalize well. Indeed, they tend to prefer certain representations of the dynamics where the symmetries and conservation laws of the system are not exactly enforced. As a result, these models do not generalize well as they are often not capable of producing physically plausible results when applied to new unseen states. Deep learning models capable of learning and generalizing dynamics effectively are typically over-parameterized, and as a consequence tend to have high variance and can be very difficult to interpret \cite{Willard2020}. Also, training these models usually requires large datasets and a long computational time, which makes them prohibitively expensive for many applications.  

A recent research direction is to consider a hybrid approach which combines knowledge of physics laws and deep learning architectures~\cite{BurbyHenon, Jin2020, Lei2020, Qin2020}. The idea is to encode physics laws and the conservation of geometric properties of the underlying systems in the design of the neural networks or in the learning process. Available physics prior knowledge can be used to construct physics-constrained neural networks with improved design and efficiency and a better generalization capacity, which take advantage of the function approximation power of neural networks to deal with incomplete knowledge.  \\

In this paper, we will consider the problem of learning dynamics for highly-oscillatory Hamiltonian systems. Examples include the Klein--Gordon equation in the weakly-relativistic regime, charged particles moving through a strong magnetic field, and the rotating inviscid Euler equations in quasi-geostrophic scaling \cite{Cotter_2004}. More generally, \emph{any} Hamiltonian system may be embedded as a normally-stable elliptic slow manifold in a nearly-periodic Hamiltonian system \cite{BurbyHi2021}. Highly-oscillatory Hamiltonian systems exhibit two basic structural properties whose interactions play a crucial role in their long-term dynamics. First is preservation of the symplectic form, as for all Hamiltonian systems. Second is timescale separation, corresponding to the relatively short timescale of oscillations compared with slower secular drifts. Coexistence of these two structural properties implies the existence of an adiabatic invariant\cite{Kruskal1962,BurbySquire_2020,BurbyHi2021,BurbyLeok2021}. Adiabatic invariants differ from true constants of motion, in particular energy invariants, which do not change at all over arbitrary time intervals. Instead adiabatic invariants are conserved with limited precision over very large time intervals. There are no learning frameworks available today that exactly preserve the two structural properties whose interplay gives rise to adiabatic invariants. This work addresses this challenge by exploiting a recently-developed theory of \emph{nearly-periodic symplectic maps}~\cite{BurbyLeok2021}, which can be thought of as discrete-time analogues of highly-oscillatory Hamiltonian systems \cite{Kruskal1962}.

As a result of being symplectic, a mapping assumes a number of special properties. In particular, symplectic mappings are closely related to Hamiltonian systems: any solution to a Hamiltonian system is a symplectic flow \cite{Poincare1899}, and any symplectic flow corresponds locally to an appropriate Hamiltonian system \cite{HaLuWa2006}. It is well-known that preserving the symplecticity of a Hamiltonian system when constructing a discrete approximation of its flow map ensures the preservation of many aspects of the dynamical system such as energy conservation, and leads to physically well-behaved discrete solutions over exponentially-long time intervals~\cite{IserlesWhyGNI,HaLuWa2006,Blanes2017,LeRe2005,Holm2009}. It is thus important to have structure-preserving neural network architectures which can learn symplectic maps and ensure that the learnt surrogate map preserves symplecticity. Many physics-informed and structure-preserving machine learning approaches have recently been proposed to learn Hamiltonian dynamics and symplectic maps \cite{BurbyHenon,Jin2020,Chen2020,Chen2021neural,Cranmer2020,Greydanus2019,Lutter2018,Zhong2020,zhong2020dissipative,Zhong2021,Saemundsson2020,Havens2021,LieFVINsExtended,Santos2022,Valperga2022,Bertalan2019,Rath2021,Offen2022,Marco2021,Mathiesen2022}. In particular, H\'enon Neural Networks (H\'enonNets)~\cite{BurbyHenon} can approximate arbitrary well any symplectic map via compositions of simple yet expressive elementary symplectic mappings called H\'enon-like mappings. In the numerical experiments conducted in this paper, H\'enonNets~\cite{BurbyHenon} will be our preferred choice of symplectic map approximator to use as building block in our framework for approximation of nearly-periodic symplectic maps, although some of the other approaches listed above for approximating symplectic mappings can be used within our framework as well.  \\

As shown by Kruskal \cite{Kruskal1962}, every nearly-periodic system, Hamiltonian or not, admits an approximate $U(1)$-symmetry, determined to leading order by the unperturbed periodic dynamics. It is well-known that a Hamiltonian system which admits a continuous family of symmetries also admits a corresponding conserved quantity. It is thus not surprising that a nearly-periodic Hamiltonian system, which admits an approximate symmetry, must also have an approximate conservation law~\cite{BurbyLeok2021}, and the approximately conserved quantity is referred to as an {adiabatic invariant}. 

\textit{Nearly-periodic maps}, first introduced by Burby et al.\cite{BurbyLeok2021}, are natural discrete-time analogues of nearly-periodic systems, and have important applications to numerical integration of nearly-periodic systems. Nearly-periodic maps may also be used as tools for structure-preserving simulation of non-canonical Hamiltonian systems on exact symplectic manifolds~\cite{BurbyLeok2021}, which have numerous applications across the physical sciences. Noncanonical Hamiltonian systems play an especially important role in modeling weakly-dissipative plasma systems~\cite{Morrison_1980,Morrison_MHD_1980,Morrison_fluid_1998,Burby_gvm_2015,Morrison_gen_beatification_2016,Morrison_neg_modes_1989,BurbyPOP2022}. Similarly to the continuous-time case, nearly-periodic maps with a Hamiltonian structure (that is symplecticity) admit an approximate symmetry and as a result also possess an adiabatic invariant~\cite{BurbyLeok2021}. The adiabatic invariants that our networks target only arise in purely Hamiltonian systems. Just like dissipation breaks the link between symmetries and conservation laws in Hamiltonian systems, dissipation also breaks the link between approximate symmetries and approximate conservation laws in Hamiltonian systems. We are not considering systems with symmetries that are broken by dissipation or some other mechanism, but rather considering systems which possess approximate symmetries. This should be contrasted with other frameworks \cite{Hernandez2021,Hernandez2023,Huang2022} which develop machine learning techniques for systems that explicitly include dissipation.    \\

We note that neural network architectures designed for multi-scale dynamics and long-time dependencies are available~\cite{unicornn}, and that many authors have introduced numerical algorithms specifically designed to efficiently step over high-frequency oscillations~\cite{chen2011energy, chen2015multi, miller2019imex}. However, the problem of developing surrogate models for dynamical systems that avoid resolving short oscillations remains open. Such surrogates would accelerate optimization algorithms that require querying the dynamics of an oscillatory system during the optimizer's ``inner loop". The network architecture presented in this article represents a first important step toward a general solution of this problem. Some of its advantages are that it aims to learn a fast surrogate model that can resolve long-time dynamics using very short time data, and that it is guaranteed to enjoy symplectic universal approximation within the class of nearly periodic maps. As developed in this paper, our method applies to dynamical systems that exhibit a single fast mode of oscillation. In particular, when initial conditions for the surrogate model are selected on the zero level set of the learned adiabatic invariant, the network automatically integrates along the slow manifold~\cite{Lorenz_1992,Lorenz_1987,Lorenz_1986,MacKay_2004,Burby_Klotz_2020}. While our network architecture generalizes in a straightforward manner to handle multiple non-resonant modes, it cannot be applied to dynamical systems that exhibit resonant surfaces.

\newpage

Note that many of the approaches listed earlier for physics-based or structure-preserving learning of Hamiltonian dynamics focus on learning the vector field associated to the continuous-time Hamiltonian system, while others learn a discrete-time symplectic approximation to the flow map of the Hamiltonian system. In many contexts, we do not need to infer the continuous-time dynamics, and only need a surrogate model which can rapidly generate accurate predictions which remain physically consistent for a long time. Learning a discrete-time approximation to the evolution or flow map, instead of learning the continuous-time vector field, allows for fast prediction and simulation without the need to integrate differential equations or use neural ODEs and adjoint techniques (which can be very expensive and can introduce additional errors due to discretization). In this paper, we will learn nearly-periodic symplectic approximations to the flow maps of nearly-periodic Hamiltonian systems, with the intention of obtaining algorithms which can generate accurate and physically-consistent simulations much faster than traditional integrators. \\

\noindent \textbf{Outline.} We first review briefly some background notions from differential geometry in Section~\ref{section: Differential Geometry}. Then, we discuss how symplectic maps can be approximated using H\'enonNets in Section~\ref{section: Approximation of Symplectic Maps}, before defining nearly-periodic systems and maps and reviewing their important properties in Section~\ref{section: Nearly-Periodic Systems and Nearly-Periodic Maps}. In Section~\ref{section: New NN Architecture}, we introduce novel neural network architectures, \emph{gyroceptrons} and \emph{symplectic gyroceptrons}, to approximate symplectic and non-symplectic nearly-periodic maps. We then show in Section~\ref{section: Numerical Confirmation of the Existence of Adiabatic Invariants} that symplectic gyroceptrons admit adiabatic invariants regardless of the values of their weights. Finally, in Section~\ref{section: Numerical Example}, we demonstrate how the proposed architecture can be used to learn surrogate maps for the nearly-periodic symplectic flow maps associated to two different systems: a nearly-periodic Hamiltonian system composed of two nonlinearly coupled oscillators (in Section~\ref{section: Nonlinearly Coupled Oscillators}), and the nearly-periodic Hamiltonian system describing the evolution of a charged particle interacting with its self-generated electromagnetic field (in Section~\ref{sec: higher-dimensional example}). \\

\section{Preliminaries}

\subsection{Differential Geometry Background}   \label{section: Differential Geometry}

\hfill 

In this paper, we reserve the symbol $M$ for a smooth manifold equipped with a smooth auxiliary Riemannian metric $g$, and $\mathcal{E}$ will always denote a vector space for the parameter $\varepsilon$. We will now briefly introduce some standard concepts from differential geometry that will be used throughout this paper (more details can be found in introductory differential geometry books \cite{McInerney2013,Lang1999,MaRa1999}). 

A smooth map $h :M_1\rightarrow M_2$ between smooth manifolds $M_1,M_2$ is a \textbf{diffeomorphism} if it is bijective with a smooth inverse. We say that $f_\varepsilon:M_1\rightarrow M_2$, $\varepsilon\in\mathcal{E}$, is a smooth $\varepsilon$-dependent mapping when the mapping $M_1\times\mathbb{R}\rightarrow M_2:(m,\varepsilon)\mapsto f_\varepsilon(m)$ is smooth. 

A \textbf{vector field} on a manifold $M$ is a map $X:M \rightarrow TM$ such that $X(m) \in T_mM$ for all $m\in M$, where $T_mM$ denotes the \textbf{tangent space} to $M$ at $m$ and  $TM = \{  (m,v) \, | \, m\in M, v \in T_mM  \}$ is the \textbf{tangent bundle} $TM$ of $M$. The vector space dual to $T_m M$ is the \textbf{cotangent space} $T_m^* M$, and the \textbf{cotangent bundle} of $M$ is $T^* M = \{  (m,p) \, | \, m\in M, p \in T^*_mM \}$. The integral curve at $m$ of a vector field $X$ is the smooth curve $c$ on $M$ such that $c(0)=m$ and $c'(t) = X(c(t))$. The \textbf{flow} of a vector field $X$ is the collection of maps $\varphi_t:M \rightarrow M$ such that $\varphi_t(m) $ is the integral curve of $X$ with initial condition $m\in M$.

A $\boldsymbol{k}$\textbf{-form} on a manifold $M$ is a map which assigns to every point $m\in M$ a skew-symmetric $k$-multilinear map on $T_mM$. Let $\alpha$ be a $k$-form and $\beta$ be a $s$-form $\beta$ on a manifold $M$. Their tensor product $\alpha \otimes \beta$ at $m\in M$ is defined via \[  (\alpha \otimes \beta)_m (v_1 , \ldots , v_{k+s} ) = \alpha_m(v_1, \ldots, v_k) \beta_m (v_{k+1} , \ldots, v_{k+s}).\]
The alternating operator $\text{Alt}$ acts on a $k$-form $\alpha$ via
\[  \text{Alt}(\alpha)(v_1 , \ldots , v_k) = \frac{1}{k!} \sum_{\pi \in S_k}{\text{sgn}(\pi) \alpha(v_{\pi(1)} , \ldots , v_{\pi(k)})}, \]
where $S_k$ is the group of all the permutations of $\{ 1, \ldots, k\}$ and $\text{sgn}(\pi)$ is the sign of the permutation. The \textbf{wedge product} $\alpha \wedge \beta$ is then defined via
\[ \alpha \wedge \beta = \frac{(k+s)!}{k!s!} \text{Alt}(\alpha \otimes \beta). \]
The \textbf{exterior derivative} of a smooth function $f: M \rightarrow \mathbb{R}$ is its differential $\mathbf{d}f$, and the \textbf{exterior derivative} $\mathbf{d}\alpha$ of a $k$-form $\alpha$ with $k>0$ is the $(k+1)$-form defined by \[  \mathbf{d} \left(  \sum_{i_1, \ldots , i_k}{ \alpha_{i_1 \ldots i_k} \mathbf{d}x^{i_1} \wedge \ldots \wedge \mathbf{d}x^{i_k}  }\right)  = \sum_{j}{ \sum_{i_1, \ldots , i_k}{ \partial_j \alpha_{i_1 \ldots i_k} \mathbf{d} x^j \wedge \mathbf{d}x^{i_1} \wedge \ldots \wedge \mathbf{d}x^{i_k}  } }  . \]	
The \textbf{interior product} $\iota_X \alpha$ where $X$ is a vector field on $M$ and $\alpha$ is a $k$-form is the $(k-1)$-form defined via \[(\iota_X \alpha)_m (v_2, \ldots , v_k) = \alpha_m(X(m), v_2, \ldots , v_k). \]
The \textbf{pull-back} $\psi^* \alpha $ of $\alpha$ by a smooth map $\psi :M \rightarrow N$ is the $k$-form defined by
\[ (\psi^* \alpha)_m(v_1,\ldots,v_k) =  \alpha_{\psi(m) } (\mathbf{d}\psi \cdot v_1 , \ldots , \mathbf{d}\psi \cdot v_k )  .\]
The \textbf{Lie derivative} $\mathcal{L}_{X} \alpha$ of the $k$-form $\alpha$ along a vector field $X$ with flow $\varphi_t$ is $\mathcal{L}_{X} \alpha = \frac{d}{dt} \Big\rvert_{t=0} \varphi_t^* \alpha $, and for a smooth function $f : M \rightarrow \mathbb{R}$, $\mathcal{L}_{X} f$ is the directional derivative $\mathcal{L}_{X} f = \mathbf{d}f \cdot X$. 

The \textbf{circle group} $U(1)$, also known as first unitary group, is the one-dimensional Lie group of complex numbers of unit modulus with the standard multiplication operation. It can be parametrized via $e^{i\theta}$ for $\theta \in [0,2\pi )$, and is isomorphic to the special orthogonal group $\text{SO}(2)$ of rotations in the plane. A \textbf{circle action} on a manifold $M$ is a one-parameter family of smooth diffeomorphisms $\Phi_\theta : M \rightarrow M$ that satisfies the following three properties for any  $\theta,\theta_1,\theta_2 \in U(1)  \cong  \mathbb{R} \text{ mod } 2\pi $:
\[   \Phi_{\theta + 2\pi} = \Phi_\theta \quad \text{(periodicity),} \qquad \Phi_{0} = \text{Id}_M \quad   \text{(identity),}  \qquad \Phi_{\theta_1 + \theta_2} = \Phi_{\theta_1} \ \circ \  \Phi_{\theta_2}  \quad \text{(additivity).} \]
The \textbf{infinitesimal generator} of a circle action $\Phi_\theta $ on $M$ is the vector field on $M$ defined by $ m \mapsto  \frac{d}{d\theta} \Big\rvert_{\theta=0} \Phi_{\theta} (m)$. \\

\subsection{Approximation of Symplectic Maps via H\'enon Neural Networks}  \label{section: Approximation of Symplectic Maps}

\hfill 

Let $U\subset \mathbb{R}^{n}\times\mathbb{R}^n=\mathbb{R}^{2n}$ be an open set in an even-dimensional Euclidean space. Denote points in $\mathbb{R}^n\times\mathbb{R}^n$ using the notation $(x,y)$, with $x,y\in\mathbb{R}^n$. A smooth mapping $\Phi:U\rightarrow\mathbb{R}^{2n}$ with components $\Phi(x,y) = (\bar{x}(x,y),\bar{y}(x,y))$ is symplectic if
\begin{align}\label{symplectic_property}
	\sum_{i=1}^n \mathbf{d}x^i\wedge \mathbf{d}y^i = \sum_{i=1}^n \mathbf{d}\bar{x}^i\wedge \mathbf{d}\bar{y}^i.
\end{align}
The symplectic condition \eqref{symplectic_property} implies that the mapping $\Phi$ has a number of special properties. In particular, there is a close relation between Hamiltonian systems and symplecticity of flows: Poincar\'e's Theorem~\cite{Poincare1899} states that any solution to a Hamiltonian system is a symplectic flow, and it can also be shown that any symplectic flow corresponds locally to an appropriate Hamiltonian system. Preserving the symplecticity of a Hamiltonian system when constructing a discrete approximation of its flow map ensures the preservation of many aspects of the dynamical system such as energy conservation, and leads to physically well-behaved discrete solutions~\cite{IserlesWhyGNI,HaLuWa2006,Blanes2017,LeRe2005,Holm2009}. It is thus important to have structure-preserving network architectures which can learn symplectic maps.

The space of all symplectic maps is infinite dimensional \cite{Weinstein1971}, so the problem of approximating an arbitrary symplectic map using compositions of simpler symplectic mappings is inherently interesting. Turaev~\cite{Turaev2002} showed that every symplectic map may be approximated arbitrarily well by compositions of \emph{H\'enon-like maps}, which are special elementary symplectic maps.
\begin{definition}
	Let $V:\mathbb{R}^n\rightarrow\mathbb{R}$ be a smooth function on $\mathbb{R}^n$ and let $\eta\in\mathbb{R}^n$ be a constant. We define the \textbf{H\'enon-like map} $H[V,\eta]:\mathbb{R}^n\times\mathbb{R}^n\rightarrow\mathbb{R}^n\times\mathbb{R}^n$ with potential $V$ and shift $\eta$ via
	\begin{align} \label{eq: Henon Map}
		H[V,\eta]\begin{pmatrix} x\\y \end{pmatrix} = \begin{pmatrix} y + \eta\\ -x +\nabla V(y) \end{pmatrix}.
	\end{align}
\end{definition} 
\begin{theorem}[Turaev \cite{Turaev2002}]\label{thm1}
	Let $\Phi:U\rightarrow\mathbb{R}^n\times\mathbb{R}^n$ be a $C^{r+1}$ symplectic mapping. For each compact set $C\subset U$ and $\delta >0$ there is a smooth function $V:\mathbb{R}^n\rightarrow\mathbb{R}$, a constant $\eta$, and a positive integer $N$ such that $H[V,\eta]^{4N}$ approximates the mapping $\Phi$ within $\delta$ in the $C^r$ topology.
\end{theorem}
\begin{remark}
	The significance of the number $4$ in this theorem follows from the fact that the fourth iterate of the H\'enon-like map with trivial potential $V=0$ is the identity map: $H[0,\eta]^4 = \emph{Id}_{\mathbb{R}^{n} \times \mathbb{R}^n}$.
\end{remark} 

Turaev's result suggests the specific neural network architecture to approximate symplectic mappings using H\'enon-like maps \cite{BurbyHenon}. We review the construction of H\'enonNets \cite{BurbyHenon}, starting with the notion of a \emph{H\'enon layer}.
\begin{definition}
	Let $\eta\in\mathbb{R}^n$ be a constant vector, and let $V$ be a scalar feed-forward neural network on~$\mathbb{R}^n$, that is., a smooth mapping $V:\mathcal{W}\times\mathbb{R}^n\rightarrow \mathbb{R}$, where $\mathcal{W}$ is a space of neural network weights. The \textbf{H\'enon layer} with potential $V$, shift $\eta$, and weight $W$ is the iterated H\'enon-like map \begin{equation} L[V[W],\eta] = H[V[W],\eta]^4,\end{equation}
	where we use the notation $V[W]$ to denote the mapping $V[W](y) = V(W,y), $ for any $ y\in\mathbb{R}^n, \text{ } W\in\mathcal{W}. $
\end{definition}

\noindent There are various network architectures for the potential $V[W]$ that are capable of approximating any smooth function $V:\mathbb{R}^n\rightarrow\mathbb{R}$ with any desired level of accuracy. For example, a fully-connected neural network with a single hidden layer of sufficient width can approximate any smooth function. Therefore a corollary of Theorem \ref{thm1} is that any symplectic map may be approximated arbitrarily well by the composition of sufficiently many H\'enon layers with various potentials and shifts. This leads to the notion of a \emph{H\'enon Neural Network}.  

\begin{definition} \label{def: HenonNet}
	Let  $N$ be a positive integer and
	\begin{itemize} \setlength{\itemindent}{8mm}
		\item  $\bm{V} = \{V_k\}_{k\in \{1,\dots, N\}}$ be a family of scalar feed-forward neural networks on $\mathbb{R}^n$ 
		\item $\bm{W} = \{W_k\}_{k\in\{1,\dots,N\}}$ be a family of network weights for $\bm{V}$
		\item $\bm{\eta} = \{\eta_k\}_{k\in\{1,\dots,N\}}$ be a family of constants in $\mathbb{R}^n$
	\end{itemize}
	The \textbf{H\'enon neural network (H\'enonNet)} with layer potentials $\bm{V}$, layer weights $\bm{W}$, and layer shifts $\bm{\eta}$ is the mapping 
	\begin{align}
		\mathcal{H}[\bm{V}[\bm{W}],\bm{\eta}] & \ =\  L[V_N[W_N],\eta_N] \ \ \circ \  \ \dots \ \ \circ \  \ L[V_2[W_2],\eta_2] \ \ \circ \  \ L[V_1[W_1],\eta_1] \\ & \ =\  H[V_N[W_N],\eta_N]^4 \ \ \circ \  \ \dots \ \ \circ \  \ H[V_2[W_2],\eta_2]^4 \ \ \circ \  \ H[V_1[W_1],\eta_1]^4.
	\end{align} 
\end{definition}

A composition of symplectic mappings is also symplectic, so every H\'enonNet is a symplectic mapping, regardless of the architectures for the networks $V_k$ and of the weights $W_k$. Furthermore, Turaev's Theorem~\ref{thm1} implies that the family of H\'enonNets is sufficiently expressive to approximate any symplectic mapping:
\begin{lemma}
	Let $\Phi:U\rightarrow\mathbb{R}^n\times\mathbb{R}^n$ be a $C^{r+1}$ symplectic mapping. For each compact set $C\subset U$ and $\delta >0$ there is a H\'enonNet $\mathcal{H}$ that approximates $\Phi$ within $\delta$ in the $C^r$ topology. 
\end{lemma}

\begin{remark} \label{remark: Henon Invertibility}
	Note that H\'enon-like maps are easily invertible,
	\begin{align} 
		H[V,\eta]\begin{pmatrix} x\\y \end{pmatrix} = \begin{pmatrix} y + \eta\\ -x +\nabla V(y) \end{pmatrix}  \quad  \Rightarrow  \quad H^{-1}[V,\eta]\begin{pmatrix} x\\y \end{pmatrix} = \begin{pmatrix} \nabla V(x-\eta) - y \\ x - \eta   \end{pmatrix} ,
	\end{align}
	so we can also easily invert H\'enon networks by composing inverses of H\'enon-like maps.
\end{remark}

We also introduce here modified versions of H\'enon-like maps and H\'enonNets to approximate symplectic maps possessing a near-identity property:
\begin{definition}
	Let $V:\mathbb{R}^n\rightarrow\mathbb{R}$ be a smooth function and let $\eta\in\mathbb{R}^n$ be a constant. We define the \textbf{near-identity H\'enon-like map} $H_\varepsilon[V,\eta]:\mathbb{R}^n\times\mathbb{R}^n\rightarrow\mathbb{R}^n\times\mathbb{R}^n$ with potential $V$ and shift $\eta$ via
	\begin{align} 
		H_\varepsilon[V,\eta]\begin{pmatrix} x\\y \end{pmatrix} = \begin{pmatrix} y + \eta\\ -x + \varepsilon \nabla V(y) \end{pmatrix}.
	\end{align}
	Near-identity H\'enon-like maps satisfy the near-identity property $H_0[V,\eta]^4 = \emph{Id}_{\mathbb{R}^n\times\mathbb{R}^n}$. 
\end{definition}

\begin{definition} \label{def: near-identity HenonNet}
	Let  $N$ be a positive integer and
	\begin{itemize} \setlength{\itemindent}{8mm}
		\item  $\bm{V} = \{V_k\}_{k\in \{1,\dots, N\}}$ be a family of scalar feed-forward neural networks on $\mathbb{R}^n$ 
		\item $\bm{W} = \{W_k\}_{k\in\{1,\dots,N\}}$ be a family of network weights for $\bm{V}$
		\item $\bm{\eta} = \{\eta_k\}_{k\in\{1,\dots,N\}}$ be a family of constants in $\mathbb{R}^n$
	\end{itemize}
	The \textbf{near-identity H\'enon network} with layer potentials $\bm{V}$, layer weights $\bm{W}$, and layer shifts $\bm{\eta}$ is the mapping defined via
	\begin{align}
		\mathcal{H}_\varepsilon[\bm{V}[\bm{W}],\bm{\eta}] \ = \
		H_\varepsilon[V_N[W_N],\eta_N]^4 \ \circ \   \ldots \ \circ \  H_\varepsilon[V_2[W_2],\eta_2]^4 \ \circ \  H_\varepsilon[V_1[W_1],\eta_1]^4,
	\end{align}
	and it satisfies the near-identity property $\mathcal{H}_0[\bm{V}[\bm{W}],\bm{\eta}] = \emph{Id}_{\mathbb{R}^n\times\mathbb{R}^n}$.  \\
\end{definition}

\subsection{Nearly-Periodic Systems and Nearly-Periodic Maps} \label{section: Nearly-Periodic Systems and Nearly-Periodic Maps}

\vspace{3mm} 

\subsubsection{Nearly-Periodic Systems}

\vspace{2.5mm}

Intuitively, a continuous-time dynamical system with parameter $\varepsilon$ is \emph{nearly-periodic} if all of its trajectories are periodic with nowhere-vanishing angular frequency in the limit $\varepsilon\rightarrow 0$. Such a system characteristically displays limiting short-timescale dynamics that ergodically cover circles in phase space. More precisely, a nearly-periodic systems can be defined as follows:

\newpage 

\begin{definition} [Burby et al. \cite{BurbyLeok2021}]
	A \textbf{nearly-periodic system} on a manifold $M$ is a smooth $\varepsilon$-dependent vector field $X_\varepsilon$ on $M$ such that $X_0 = \omega_0\,R_0$, where
	\begin{itemize} \setlength{\itemindent}{8mm}
		\item $R_0$ is the infinitesimal generator for a circle action $\Phi_\theta:M\rightarrow M$, $\theta\in U(1)$.
		\item $\omega_0:M\rightarrow\mathbb{R}$ is strictly positive and its Lie derivative satisfies $\mathcal{L}_{R_0}\omega_0 = 0$.
	\end{itemize}
	The vector field $R_0$ is called the \textbf{limiting roto-rate}, and $\omega_0$ is the \textbf{limiting angular frequency}.
\end{definition}

Examples from physics include charged particle dynamics in a strong magnetic field, the weakly-relativistic Dirac equation, and any mechanical system subject to a high-frequency, time-periodic force. In the broader context of multi-scale dynamical systems, nearly-periodic systems play a special role because they display perhaps the simplest possible non-dissipative short-timescale dynamics. They therefore provide a useful proving ground for analytical and numerical methods aimed at more complex multi-scale models.
\begin{remark}
	In a paper \cite{Kruskal1962} on basic properties of continuous-time nearly-periodic systems, Kruskal assumed that $R_0$ is nowhere vanishing, in addition to requiring that $\omega_0$ is sign-definite. This assumption is usually not essential and it is enough to require that $\omega_0$ vanishes nowhere. This is an important restriction to lift since many interesting circle actions have fixed points. 
\end{remark} 

It can be shown that every nearly-periodic system admits an approximate $U(1)$-symmetry \cite{Kruskal1962}, known as the \emph{roto-rate}, that is determined to leading order by the unperturbed periodic dynamics:
\begin{definition}
	A \textbf{roto-rate} for a nearly-periodic system $X_\varepsilon$ on a manifold $M$ is a formal power series $R_\varepsilon = R_0 + \varepsilon\,R_1 + \varepsilon^2\,R_2 + \dots$ with vector field coefficients such that $R_0$ is equal to the limiting roto-rate and the following equalities hold in the sense of formal series:
	\[ \exp(2\pi \mathcal{L}_{R_\varepsilon}) = 1 \qquad \text{and} \qquad [X_\varepsilon,R_\varepsilon] = 0 .\] 
\end{definition}

\begin{proposition}[Kruskal \cite{Kruskal1962}]\label{existence_of_roto_rate}
	Every nearly-periodic system admits a unique roto-rate $R_\varepsilon$. 
\end{proposition}

A subtle argument allows to upgrade leading-order $U(1)$-invariance to all-orders $U(1)$-invariance for integral invariants:
\begin{proposition}[Burby et al.\cite{BurbyLeok2021}]\label{bootstrap_prop}
	Let $\alpha_\varepsilon$ be a smooth $\varepsilon$-dependent differential form on a manifold $M$. Suppose $\alpha_\varepsilon$ is an absolute integral invariant for a smooth nearly-periodic system $X_\varepsilon$ on $M$. If $\mathcal{L}_{R_0}\alpha_0 = 0$ then $\mathcal{L}_{R_\varepsilon}\alpha_\varepsilon = 0$, where $R_\varepsilon$ is the roto-rate for $X_\varepsilon$. \\
\end{proposition}

\subsubsection{Nearly-Periodic Maps}

\hfill 

Nearly-periodic maps are natural discrete-time analogues of nearly-periodic systems, which were first introduced in \cite{BurbyLeok2021}. The following provides a precise definition.
\begin{definition}\label{def: np map}
	A \textbf{nearly-periodic map} on a manifold $M$ with parameter vector space $\mathcal{E}$ is a smooth mapping $F:M\times \mathcal{E}\rightarrow M$ such that $F_\varepsilon:M\rightarrow M:m\mapsto F(m,\varepsilon)$ has the following properties:
	\begin{itemize} \setlength{\itemindent}{8mm}
		\item $F_\varepsilon$ is a diffeomorphism for each $\varepsilon\in \mathcal{E}$.
		\item There exists a $U(1)$-action $\Phi_\theta:M\rightarrow M$ and a constant $\theta_0\in U(1)$ such that $F_0 = \Phi_{\theta_0}$.
	\end{itemize}
	We say $F$ is \textbf{resonant} if $\theta_0$ is a rational multiple of $2\pi$, otherwise $F$ is \textbf{non-resonant}. The infinitesimal generator of $\Phi_\theta$, $R_0$, is the \textbf{limiting roto-rate}.
\end{definition}

\begin{example}
	Let $X_\varepsilon$ be a nearly-periodic system on a manifold $M$ with limiting roto-rate $R_0$ and limiting angular frequency $\omega_0$. Assume that $\omega_0$ is constant. For each $\varepsilon\in \mathbb{R}$ let $\mathcal{F}_t^\varepsilon$ denote the time-$t$ flow for $X_\varepsilon$. The mapping $F(m,\varepsilon) = \mathcal{F}_{t_0}^\varepsilon(m) $ is nearly-periodic for each $t_0$. To see why, first note that the flow of the limiting vector field $X_0 = \omega_0\,R_0$ is given by $\mathcal{F}_t^0(m) = \Phi_{\omega_0\,t}(m)$, where $\Phi_\theta$ denotes the $U(1)$-action generated by $R_0$. It follows that $	F(m,0) = \Phi_{\omega_0\,t_0}(m) = \Phi_{\theta_0}(m)$, where $\theta_0 = \omega_0\,t_0\emph{ mod }2\pi$. This example is more general than it first appears since any nearly-periodic system can be rescaled to have a constant limiting angular frequency. Indeed if the nearly-periodic system $X_\varepsilon$ has non-constant limiting angular frequency $\omega_0$ then $X^\prime_\varepsilon = X_\varepsilon/\omega_0$ is a nearly-periodic system with limiting angular frequency $1$. The integral curves of $X^\prime_\varepsilon$ are merely time reparameterizations of integrals curves of $X_\varepsilon$.
\end{example}

Let $X$ be a vector field on a manifold $M$ with time-$t$ flow map $\mathcal{F}_t$. A $U(1)$-action $\Phi_\theta$ is a \emph{$U(1)$-symmetry} for $X$ if $\mathcal{F}_t\ \circ \  \Phi_\theta = \Phi_\theta\ \circ \  \mathcal{F}_t$, for each $t\in\mathbb{R}$ and $\theta\in U(1)$. Differentiating this condition with respect to~$\theta$ at the identity implies, and is implied by, $\mathcal{F}_t^*R = R$, where $R$ denotes the infinitesimal generator for the $U(1)$-action. Since we would like to think of nearly-periodic maps as playing the part of a nearly-periodic system's flow map, the latter characterization of symmetry allows us to naturally extend Kruskal's notion of roto-rate to our discrete-time setting.
\begin{definition}
	A \textbf{roto-rate} for a nearly-periodic map $F:M\times \mathcal{E}\rightarrow M$ is a formal power series $R_\varepsilon = R_0 + R_1\varepsilon + R_2\varepsilon^2+\dots$ whose coefficients are vector fields on $M$ such that $R_0$ is the limiting roto-rate and the following equalities hold in the sense of formal power series: $F_\varepsilon^*R_\varepsilon = R_\varepsilon $ and $ \exp(2\pi\mathcal{L}_{R_\varepsilon})=1$.
\end{definition}

A first fundamental result concerning nearly-periodic maps establishes the existence and uniqueness of the roto-rate in the non-resonant case. Like the corresponding result in continuous time, this result holds to all orders in perturbation theory. 

\begin{theorem}[Burby et al.\cite{BurbyLeok2021}] 
	Each non-resonant nearly-periodic map admits a unique roto-rate. 
\end{theorem}
Thus, non-resonant nearly-periodic maps formally reduce to mappings on the space of $U(1)$-orbits, corresponding to the elimination of a single dimension in phase space. \\

\subsubsection{Nearly-Periodic Systems and Maps with a Hamiltonian Structure}

\vspace{3mm}

\begin{definition}
	A \textbf{$\varepsilon$-dependent presymplectic manifold} is a manifold $M$ equipped with a smooth $\varepsilon$-dependent $2$-form $\Omega_\varepsilon$ such that $\mathbf{d}\Omega_\varepsilon = 0$ for each $\varepsilon\in \mathcal{E}$. We say $(M,\Omega_\varepsilon)$ is \textbf{exact} when there is a smooth $\varepsilon$-dependent $1$-form $\vartheta_\varepsilon$ such that $\Omega_\varepsilon = -\mathbf{d}\vartheta_\varepsilon$.
\end{definition}
\begin{definition}\label{nearly_periodic_hamiltonian_system_def}
	A \textbf{nearly-periodic Hamiltonian system} on an exact presymplectic manifold $(M,\Omega_\varepsilon)$ is a nearly-periodic system $X_\varepsilon$ on $M$ such that $\iota_{{X}_\varepsilon}\Omega_\varepsilon = \mathbf{d}H_\varepsilon$, for some smooth $\varepsilon$-dependent function $H_\varepsilon:M\rightarrow\mathbb{R}$. 
\end{definition}

We already know from Proposition \ref{existence_of_roto_rate} that every nearly-periodic system admits a unique roto-rate~$R_\varepsilon$. In the Hamiltonian setting, it can be shown that both the dynamics and the Hamiltonian structure are $U(1)$-invariant to all orders in $\varepsilon$. 
\begin{proposition}[Kruskal\cite{Kruskal1962}, Burby et al.\cite{BurbyLeok2021}]
	The roto-rate $R_\varepsilon$ for a nearly-periodic Hamiltonian system $X_\varepsilon$ on an exact presymplectic manifold $(M,\Omega_\varepsilon)$ with Hamiltonian $H_\varepsilon$ satisfies $\mathcal{L}_{R_\varepsilon}H_\varepsilon = 0$, and $\mathcal{L}_{R_\varepsilon}\Omega_\varepsilon = 0$ in the sense of formal power series. 
\end{proposition}

According to Noether's celebrated theorem, a Hamiltonian system that admits a continuous family of symmetries also admits a corresponding conserved quantity \cite{MaRa1999,AbMa1978,Ar1989}. Therefore one might expect that a Hamiltonian system with an approximate symmetry must also have an approximate conservation law. This is indeed the case for nearly-periodic Hamiltonian systems:
\begin{proposition}[Burby et al.\cite{BurbyLeok2021}]\label{existence_of_mu}
	Let $X_\varepsilon$ be a nearly-periodic Hamiltonian system on the exact presymplectic manifold $(M,\Omega_\varepsilon)$. Let $R_\varepsilon$ be the associated roto-rate. There is a formal power series $\theta_\varepsilon = \theta_0 + \varepsilon\,\theta_1 + \dots$ with coefficients in $\Omega^1(M)$ such that $\Omega_\varepsilon = -\mathbf{d}\theta_\varepsilon$ and $\mathcal{L}_{R_\varepsilon}\theta_\varepsilon = 0$. Moreover, the formal power series $\mu_\varepsilon = \iota_{R_\varepsilon}\theta_\varepsilon$ is a constant of motion for $X_\varepsilon$ to all orders in perturbation theory. In other words,
	$
	\mathcal{L}_{X_\varepsilon}\mu_\varepsilon = 0,
	$
	in the sense of formal power series. The formal constant of motion $\mu_\varepsilon$ is the \textbf{adiabatic invariant} associated with the nearly-periodic Hamiltonian system.
\end{proposition}

Note that general expressions for the adiabatic invariant $\mu_\varepsilon$ can be obtained \cite{BurbySquire2020}. It can also be shown that the (formal) set of fixed points for the roto-rate is an elliptic almost invariant slow manifold whose normal stability is mediated by the adiabatic invariant associated with the nearly-periodic Hamiltonian system~\cite{BurbyHi2021}. \\ 

A similar theory can be established for nearly-periodic maps with a Hamiltonian structure.

\begin{definition}
	A \textbf{presymplectic nearly-periodic map} on a $\varepsilon$-dependent presymplectic manifold $(M,\Omega_\varepsilon)$ is a nearly-periodic map $F$ such that $F_\varepsilon^*\Omega_\varepsilon = \Omega_\varepsilon$ for each $\varepsilon\in \mathcal{E}$.
\end{definition}
\begin{theorem}[Burby et al.\cite{BurbyLeok2021}]
	If $F$ is a non-resonant presymplectic nearly-periodic map on a $\varepsilon$-dependent presymplectic manifold $(M,\Omega_\varepsilon)$ with roto-rate $R_\varepsilon$ then $\mathcal{L}_{R_\varepsilon}\Omega_\varepsilon = 0$.
\end{theorem}

\begin{definition}
	A \textbf{Hamiltonian nearly-periodic map} on a $\varepsilon$-dependent presymplectic manifold $(M,\Omega_\varepsilon)$ is a nearly-periodic map $F$ such that there is a smooth $(t,\varepsilon)$-dependent vector field $Y_{t,\varepsilon}$ with $t\in \mathbb{R} $ such that the following properties hold true:
	\begin{itemize} \setlength{\itemindent}{8mm}
		\item $\iota_{Y_{t,\varepsilon}}\Omega_\varepsilon = \mathbf{d}H_{t,\varepsilon}$, for some smooth $(t,\varepsilon)$-dependent function $H_{t,\varepsilon}$.
		\item For each $\varepsilon\in\mathcal{E}$, $F_\varepsilon$ is the $t=1$ flow of $Y_{t,\varepsilon}$.
	\end{itemize} 
\end{definition}

\begin{lemma}
	Each Hamiltonian nearly-periodic map is a presymplectic nearly-periodic map.
\end{lemma}

Using presymplecticity of the roto-rate, Noether's theorem can be used to establish existence of adiabatic invariants for many interesting presymplectic nearly-periodic maps.
\begin{theorem}[Burby et al. \cite{BurbyLeok2021}]
	Let $F$ be a non-resonant presymplectic nearly-periodic map on the exact $\varepsilon$-dependent presymplectic manifold $(M,\Omega_\varepsilon)$ with roto-rate $R_\varepsilon$. Assume that $F$ is Hamiltonian
	or that the manifold $M$ is connected and the limiting roto rate $R_0$ has at least one zero.
	Then there exists a smooth $\varepsilon$-dependent $1$-form $\theta_\varepsilon$ such that $\mathcal{L}_{R_\varepsilon}\theta_\varepsilon = 0$ and $-\mathbf{d}\theta_\varepsilon =\Omega_\varepsilon$ in the sense of formal power series. Moreover the quantity
	$ \mu_\varepsilon = \iota_{R_\varepsilon}\theta_\varepsilon\label{the_adiabatic_invariant}$
	satisfies $F_\varepsilon^*\mu_\varepsilon = \mu_\varepsilon$ in the sense of formal power series, that is,   $\mu_\varepsilon$ is an adiabatic invariant for $F$.
\end{theorem}

When an adiabatic invariant exists, the phase-space dimension is formally reduced by two. On the slow manifold $\mu_\varepsilon = 0$ the reduction in dimensionality may be even more dramatic. For example, the slow manifold for the symplectic Lorentz system~\cite{BurbyHi2021} has half the dimension of the full system. \\

\section{Novel Structure-Preserving Neural Network Architectures} \label{section: New NN Architecture}

\subsection{Approximating Nearly-Periodic Maps via Gyroceptrons}

\hfill 

We first consider the problem of approximating an arbitrary nearly-periodic map $P:M\times \mathcal{E}\rightarrow M$
on a manifold $M$. From Definition~\ref{def: np map}, there must be a corresponding circle action $\Phi_{\theta} : M \rightarrow M$ and $\theta_0 \in U(1)$ such that $P_0 = \Phi_{\theta_0} $. Consider the map $I_\varepsilon : M \rightarrow M$ given by 
\begin{equation} \label{eq: near-identity map}
	I_{\varepsilon} \ = \  P_\varepsilon \ \circ \   \Phi_{\theta_0}^{-1} \qquad \forall \varepsilon \in \mathcal{E}.
\end{equation} 
This defines a near-identity map on $M$ satisfying $I_0 = \text{Id}_M $. By composing both sides of equation~\eqref{eq: near-identity map} on the right by $\Phi_{\theta_0}$, we obtain a representation for any nearly-periodic map $P$ as the composition of a near-identity map and a circle action,
\begin{equation}
	P_\varepsilon \ = \  I_\varepsilon \ \circ \   \Phi_{\theta_0}  \qquad \forall \varepsilon \in \mathcal{E}.
\end{equation}

As a consequence, if we can approximate any near-identity map and any circle action, then by the above representation we can approximate any nearly-periodic map.  

Different circle actions can act on manifolds in topologically different ways, so it would be very challenging, if not impossible, to construct a single strategy which allows to approximate any circle action to arbitrary accuracy. Here, we will consider the simpler case where we assume that we know a priori the topological type of action for the nearly-periodic system, and work within conjugation classes. Conjugation of a circle action $\Phi_{\theta} : M \rightarrow M$ with a diffeomorphism $\psi$ results in the map $\psi \ \circ \  \Phi_{\theta} \ \circ \  \psi^{-1}$, and two circle actions belong to the same conjugation class if one can be written as the conjugation with a diffeomorphism of the other one. Note that although compositions of nearly-periodic maps are not necessarily nearly-periodic, the map obtained by conjugation of a nearly-periodic map with a diffeomorphism is nearly-periodic: \begin{lemma} \label{lemma: np conjugation} 
	Let $P:M\times \mathcal{E}\rightarrow M$ be a nearly-periodic map
	on a manifold $M$, and let $\psi: M \rightarrow M$ be a diffeomorphism on $M$. Then the map $\tilde{P}: M\times \mathcal{E}\rightarrow M$ defined for any $\varepsilon \in \mathcal{E}$  via
	\begin{equation} \tilde{P}_\varepsilon   \equiv \  \psi \ \circ \  P_\varepsilon \ \circ \  \psi^{-1} \end{equation} 	is a nearly-periodic map.  \\
	\textbf{Proof.} $\psi$ and $P_\varepsilon$ are diffeomorphisms for any $\varepsilon \in \mathcal{E}$ so $\tilde{P}_\varepsilon$ is also a diffeomorphism for any $\varepsilon \in \mathcal{E}$. Now, from Definition~\ref{def: np map}, there is a circle action $\Phi_{\theta} : M \rightarrow M$ and $\theta_0 \in U(1)$ such that $P_0 = \Phi_{\theta_0} $. Define  $\tilde{\Phi}_\theta : M \rightarrow M$ via $\tilde{\Phi}_\theta   \equiv \  \psi \ \circ \  \Phi_{\theta} \ \circ \  \psi^{-1} $ for any $\theta \in U(1)$. Then, for any $\theta,\theta_1,\theta_2 \in U(1)$,
		\begin{itemize} \setlength{\itemindent}{6mm}
			\item $\tilde{\Phi}_{\theta+2\pi}  \ = \ \psi \ \circ \  \Phi_{\theta+2\pi} \ \circ \  \psi^{-1}  \ = \ \psi \ \circ \  \Phi_{\theta} \ \circ \  \psi^{-1} \ = \ \tilde{\Phi}_\theta $
			\item  $\tilde{\Phi}_{0}  \ = \ \psi \ \circ \  \Phi_{0} \ \circ \  \psi^{-1}  \ = \ \psi \ \circ \  \emph{Id}_M \ \circ \  \psi^{-1}  \ = \ \emph{Id}_M $
			\item $\tilde{\Phi}_{\theta_1} \ \circ \  \tilde{\Phi}_{\theta_2} \ = \ \psi \ \circ \  \Phi_{\theta_1} \ \circ \  \psi^{-1} \ \circ \  \psi \ \circ \  \Phi_{\theta_2} \ \circ \  \psi^{-1}  \ = \ \psi \ \circ \  \Phi_{\theta_1} \ \circ \   \Phi_{\theta_2} \ \circ \  \psi^{-1}  \ = \ \psi \ \circ \  \Phi_{\theta_1+\theta_2}  \ \circ \  \psi^{-1} \ = \ \tilde{\Phi}_{\theta_1+\theta_2}  $
		\end{itemize}
		Therefore, $\tilde{\Phi}_\theta$ is a circle action, and $\theta_0 \in U(1)$ is such that $ \ \tilde{\Phi}_{\theta_0}  \ = \  \psi \ \circ \  \Phi_{\theta_0} \ \circ \  \psi^{-1} \ = \  \psi \ \circ \  P_0 \ \circ \  \psi^{-1} \ = \ \tilde{P}_0$. \\ As a consequence, $\tilde{P}$ is a nearly-periodic map.  \hfill $\blacksquare $
\end{lemma}

We also have the following useful factorization result for nearly-periodic maps with limiting rotation within a given conjugacy class:
\begin{lemma}
	Let $\Phi_\theta:M\rightarrow M$ be a circle action on a manifold $M$. Every nearly-periodic map $P_\varepsilon:M\rightarrow M$ whose limiting rotation $\Phi^\prime_{\theta_0} = P_0$ is conjugate to $\Phi_{\theta_0}$ admits the decomposition
	\begin{align}
		P_\varepsilon \ = \  I_\varepsilon\ \circ \  \psi\ \circ \ \Phi_{\theta_0}\ \circ \  \psi^{-1},
	\end{align}
	where $\psi:M\rightarrow M$ is a diffeomorphism and $I_\varepsilon:M\rightarrow M$ is a near-identity diffeomorphism.
\end{lemma}

We will thus assume that we know in advance the topological type of the circle action $\Phi_{\theta}$ for the dynamics of interest, and then propose to learn the nearly-periodic map $P_\varepsilon$ by learning each component map in the composition
\begin{equation}   P_\varepsilon \ = \  I_\varepsilon \ \circ \  \psi \ \circ \  \Phi_{\theta} \ \circ \  \psi^{-1}. \end{equation} This formula may be interpreted intuitively as follows. The map $\psi$ learns the mode structure of an oscillatory system's short timescale dynamics. The circle action $\Phi_\theta$ provides an aliased phase advance for the learnt mode. Finally, $I_\varepsilon$ captures the averaged dynamics that occurs on timescales much larger than the limiting oscillation period.

$I_\varepsilon$ and $\psi$ can be learnt using any standard neural network architecture, as long as the near-identity property is enforced in the representation for $I_\varepsilon$. It is however important to invert $\psi$ exactly, and this strongly motivates using explicitly invertible neural network architectures for $\psi$. It has been shown that those coupling-based invertible neural networks are universal diffeomorphism approximators \cite{Teshima2020}. The parameter $\theta$ in the circle action $\Phi_\theta$ can also be considered as a trainable parameter. We will refer to the resulting architecture as a  {\textbf{gyroceptron}}, named after a combination of gyrations of phase with perceptron.
\begin{definition}
	A \textbf{gyroceptron} is a feed-forward neural network \begin{equation} P_\varepsilon[W] \ = \  I_\varepsilon[W_I] \ \circ \  \psi[W_\psi] \ \circ \  \Phi_{\theta} \ \circ \  \psi[W_\psi]^{-1} \end{equation}  with weights $W=(W_I,W_\psi)$ and rotation parameter $\theta\in U(1)$, where 
	\begin{itemize} \setlength{\itemindent}{8mm}
		\item $I_\varepsilon[W_I]:M\rightarrow M$ is a diffeomorphism for each $(\varepsilon,W_I)$ such that $I_0[W_I] = \emph{Id}_M$ for each $W_I$
		\item $\psi[W_\psi]:M\rightarrow M$ is a diffeomorphism for each $W_\psi$
		\item $\Phi_\theta:M\rightarrow M$ is a circle action on $M$
	\end{itemize}
\end{definition}
Gyroceptrons enjoy the following universal approximation property.
\begin{theorem}
	Fix a circle action $\Phi_\theta:M\rightarrow M$ and a compact set $C\subset M$. Let $P_\varepsilon:M\rightarrow M$ be a nearly-periodic map whose limiting rotation is conjugate to $\Phi_\theta$. Let $\psi[W_\psi]:M\rightarrow M$ be a feed-forward network architecture that provides a universal approximation within the class of diffeomorphisms, and let $I_\varepsilon[W_I]$ be a feed-forward network architecture that provides a universal approximation within the class of $\varepsilon$-dependent diffeomorphisms with $I_0[W] = \emph{Id}_M$. For each $\delta>0$, there exist weights $W_\psi^*$ and $W_I^*$ such that  the gyroceptron $P_\varepsilon[W^*] \ = \  I_\varepsilon[W_I^*]\ \circ \  \psi[W_\psi^*]\ \circ \  \Phi_\theta\ \circ \ \psi[W_\psi^*]^{-1} $ approximates $P_\varepsilon$ within $\delta$ on $C$.   \\
\end{theorem}

\subsection{Approximating Nearly-Periodic Symplectic Maps via Symplectic Gyroceptrons} \label{section: NN for NP Symplectic}

\hfill 

We now focus on approximating an arbitrary nearly-periodic symplectic map $P:M\times \mathcal{E}\rightarrow M$
on a manifold $M$. We will restrict our attention to symplectic manifolds with $\varepsilon$-independent symplectic forms (the $\varepsilon$-dependent case is more subtle and will not be pursued in the current study). From Definition~\ref{def: np map}, there must be a corresponding symplectic circle action $\Phi_{\theta} : M \rightarrow M$ and $\theta_0 \in U(1)$ such that $P_0 = \Phi_{\theta_0} $. As before, consider the map $I_\varepsilon : M \rightarrow M$ given by 
\begin{equation} \label{eq: near-identity symplectic map}
	I_{\varepsilon} \ = \ P_\varepsilon \ \circ \   \Phi_{\theta_0}^{-1}, \qquad \forall \varepsilon \in \mathcal{E}.
\end{equation} 
Now, the inverse of a symplectic map is symplectic and any composition of symplectic maps is also symplectic. Thus, the map $\Phi_{\theta_0}^{-1} = P_0^{-1}$ is symplectic, and as a result, $ I_{\varepsilon} $ is symplectic on $M$ for any $\varepsilon \in \mathcal{E}$ and it satisfies the near-identity property $I_0 = \text{Id}_M $. By composing both sides of equation~\eqref{eq: near-identity symplectic map} on the right by $\Phi_{\theta_0}$, we obtain a representation for any nearly-periodic symplectic map~$P$ as the composition of a near-identity symplectic map and a symplectic circle action:
\begin{align}
	P_\varepsilon \ = \  I_\varepsilon \ \circ \   \Phi_{\theta_0},  \qquad \forall \varepsilon \in \mathcal{E}. \end{align}
\begin{lemma}
	\vspace{-2mm}
	Let $\Phi_\theta:M\rightarrow M$ be a symplectic  circle action on a symplectic manifold $(M,\omega)$. Every nearly-periodic symplectic map $P_\varepsilon:M\rightarrow M$ whose limiting rotation $\Phi^\prime_{\theta_0} = P_0$ is conjugate to $\Phi_{\theta_0}$ admits the decomposition
	\begin{align}
		P_\varepsilon \ = \  I_\varepsilon\ \circ \  \psi\ \circ \ \Phi_{\theta_0}\ \circ \  \psi^{-1},
	\end{align}
	where $\psi:M\rightarrow M$ is a symplectic diffeomorphism and $I_\varepsilon:M\rightarrow M$ is a near-identity symplectic diffeomorphism.
\end{lemma}

If we can approximate any near-identity symplectic map and any symplectic circle action, then by the above representation we can approximate any nearly-periodic symplectic map. As before, we will assume that we know a priori the topological type of the circle action $\Phi_{\theta}$ for the nearly-periodic symplectic system of interest, and work within conjugation classes. Since compositions of symplectic maps are symplectic, Lemma~\ref{lemma: np conjugation} implies that the map $\psi \ \circ \  P \ \circ \  \psi^{-1}$, obtained by conjugating a nearly-periodic symplectic map $P$ with a symplectomorphism $\psi$ (i.e. a symplectic diffeomorphism), is also a nearly-periodic symplectic map. We will then learn the nearly-periodic symplectic map by learning each component map in the composition 
\begin{equation}
	P_\varepsilon  \ = \  I_\varepsilon \ \circ \  \psi \ \circ \  \Phi_{\theta} \ \circ \  \psi^{-1},
\end{equation}   
where $I_\varepsilon$ is a near-identity symplectic map and $\psi$ is symplectic. 

The symplectic map $\psi$ can be learnt using any neural network architecture which strongly enforces symplecticity. It is preferable however to choose an architecture which can easily be inverted, so that the computations involving $\psi^{-1}$ can be conducted efficiently. The near-identity symplectic map $I_\varepsilon$ can be learnt using any neural network architecture strongly enforcing symplecticity with the additional property that it limits to the identity as $\varepsilon$ goes to 0. The parameter $\theta$ in the circle action $\Phi_\theta$ can also be considered as a trainable parameter. We will refer to any such resulting composition of neural network architectures as a \textbf{symplectic gyroceptron}.

\begin{definition} \label{def: symplectic gyroceptron}
	A \textbf{symplectic gyroceptron} is a feed-forward neural network \begin{equation} P_\varepsilon[W] \ = \  I_\varepsilon[W_I] \ \circ \  \psi[W_\psi] \ \circ \  \Phi_{\theta} \ \circ \  \psi[W_\psi]^{-1} \end{equation} with weights $W=(W_I,W_\psi)$ and rotation parameter $\theta\in U(1)$, where 
	\begin{itemize} \setlength{\itemindent}{8mm}
		\item $I_\varepsilon[W_I]:M\rightarrow M$ is a symplectic diffeomorphism for each $(\varepsilon,W_I)$ such that $I_0[W_I] = \emph{Id}_M$ for each $W_I$
		\item $\psi[W_\psi]:M\rightarrow M$ is a symplectic diffeomorphism for each $W_\psi$
		\item $\Phi_\theta:M\rightarrow M$ is a symplectic circle action on $M$
	\end{itemize}
\end{definition} 

Symplectic gyroceptrons enjoy a universal approximation property comparable to the non-symplectic case.
\begin{theorem}
	Fix a symplectic circle action $\Phi_\theta:M\rightarrow M$ on the symplectic manifold $(M,\omega)$ and a compact set $C\subset M$. Let $P_\varepsilon:M\rightarrow M$ be a nearly-periodic symplectic map whose limiting rotation is conjugate to $\Phi_\theta$. Let $\psi[W_\psi]:M\rightarrow M$ be a feed-forward network architecture that provides a universal approximation within the class of symplectic diffeomorphisms, and let $I_\varepsilon[W_I]$ be a feed-forward network architecture that provides a universal approximation within the class of $\varepsilon$-dependent symplectic diffeomorphisms with $I_0[W] = \emph{Id}_M$. For each $\delta>0$, there exist weights $W_\psi^*$ and $W_I^*$ such that  the symplectic gyroceptron $P_\varepsilon[W^*]  \ = \  I_\varepsilon[W_I^*]\ \circ \  \psi[W_\psi^*]\ \circ \  \Phi_\theta\ \circ \ \psi[W_\psi^*]^{-1} $ approximates $P_\varepsilon$ within $\delta$ on $C$. 
\end{theorem}

In this paper, we will use H\'enonNets~\cite{BurbyHenon} as the main building blocks of our symplectic gyroceptrons. The symplectic map $\psi$ will be learnt using a standard H\'enonNet (see Definition~\ref{def: HenonNet}), its inverse $\psi^{-1}$ can be obtained easily by composing inverses of H\'enon-like maps (see Remark~\ref{remark: Henon Invertibility}), and the near-identity symplectic map $I_\varepsilon$ will be learnt using a near-identity H\'enonNet (see Definition~\ref{def: near-identity HenonNet}). The neural network architectures considered in this paper are summarized in Figure~\ref{fig:diagram}.  \\

We would like to emphasize that symplectic building blocks other than H\'enonNets could have been used as the basis for our symplectic gyroceptrons. For instance, a possible option would have been to use SympNets~\cite{Jin2020} since they also strongly ensure symplecticity and enjoy a universal approximation property for symplectic maps. However, numerical experiments conducted in the original H\'enonNet paper~\cite{BurbyHenon} suggested that H\'enonNets have a higher per layer expressive power than SympNets, and as a result SympNets are typically much deeper than H\'enonNets, and slower for prediction. This is consistent with the observations we will make later in Section~\ref{section: Nonlinearly Coupled Oscillators} where we will see that a SympNet took 127 seconds to generate trajectories that were generated by a H\'enonNet of similar size in 3 seconds. Together with the fact that SympNets are not as easily invertible as H\'enonNets, the computational advantage of H\'enonNets makes them more desirable as building blocks than SympNets. \\ 

{
\def\layersep{2.5cm}
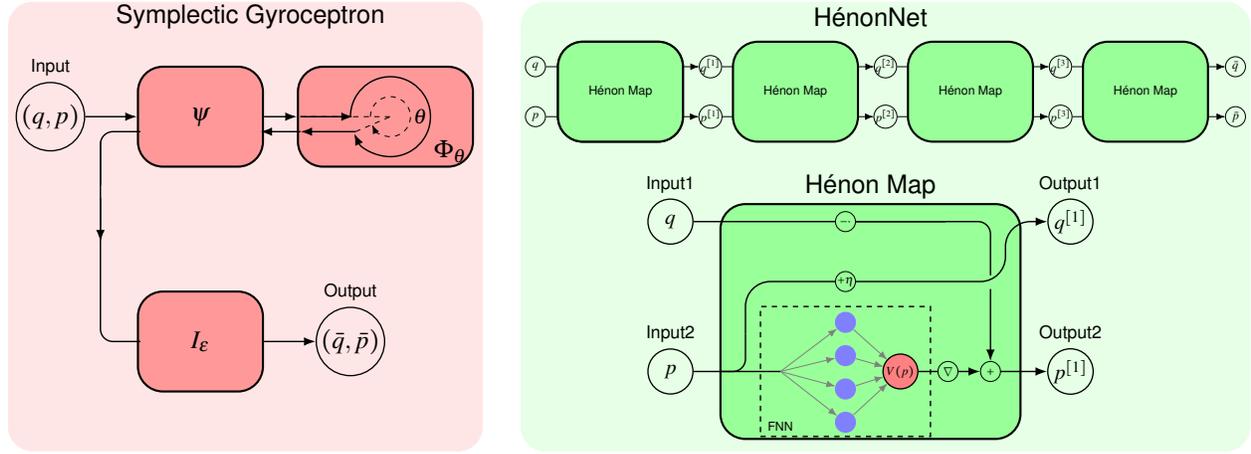
\begin{figure}[htb]
\begin{center}
\resizebox{20cm}{!}{
\begin{tikzpicture}[    
    font=\sf \scriptsize,
    >=LaTeX,
    cell/.style={
        rectangle, 
        rounded corners=5mm, 
        draw,
        very thick,
        },
    operator/.style={
        circle,
        draw,
        inner sep=-0.5pt,
        minimum height =.4cm,
        },
    function/.style={
        ellipse,
        draw,
        inner sep=1pt
        },
    ct/.style={
        circle,
        draw,
        line width = .75pt,
        minimum width=.9cm,
        inner sep=1pt,
        },
    gt/.style={
        rectangle,
        draw,
        minimum width=4mm,
        minimum height=3mm,
        inner sep=1pt
        },
    mylabel/.style={
        font=\normalsize\sffamily
        },
    ArrowC1/.style={
        rounded corners=.25cm,
        thick,
        },
    ArrowC2/.style={
        rounded corners=.5cm,
        thick,
        },]
\tikzstyle{backgroundblue}=[
    rectangle,
    fill=blue!10,
    inner sep=0.2cm,
    rounded corners=5mm
]
\tikzstyle{backgroundred}=[
    rectangle,
    fill=red!10,
    inner sep=0.2cm,
    rounded corners=5mm
]
\tikzstyle{backgroundgreen}=[
    rectangle,
    fill=green!10,
    inner sep=0.2cm,
    rounded corners=5mm
]

\useasboundingbox (0,0) rectangle (30,8); 
 
\begin{scope}[xshift=12.8cm,yshift=6.5cm]
\node [backgroundgreen, minimum height =9.cm, minimum width=14.6cm] at (5.3,-2.7){} ;
 \node [] at (5.,1.5) {\Large H\'enonNet};
 \node [operator] (in1) at (-1.7,.5) {$q$};
 \node [operator] (in2) at (-1.7,-.5) {$p$};
 \node [cell, fill=green!40, minimum height =2.cm, minimum width=2.5cm] at (0,0){H\'enon Map} ;

 \node [operator] (in11) at (1.8,.5) {$q^{[1]}$};
 \node [operator] (in12) at (1.8,-.5) {$p^{[1]}$};

 \node [operator] (in21) at (5.3,.5) {$q^{[2]}$};
 \node [operator] (in22) at (5.3,-.5) {$p^{[2]}$};

 \node [operator] (in31) at (8.8,.5) {$q^{[3]}$};
 \node [operator] (in32) at (8.8,-.5) {$p^{[3]}$};
  \node [operator] (ou1) at (12.3,.5) {$\bar q$};
 \node [operator] (ou2) at (12.3,-.5) {$\bar p$};
 
  \draw [->, draw] (in1)-- (in11);
   \draw [->, draw] (in2)-- (in12);
     \draw [->, draw] (in11)-- (in21);
   \draw [->, draw] (in12)-- (in22);
        \draw [->, draw] (in21)-- (in31);
   \draw [->, draw] (in22)-- (in32);
           \draw [->, draw] (in31)-- (ou1);
   \draw [->, draw] (in32)-- (ou2);
   
\node [cell, fill=green!40, minimum height =2.cm, minimum width=2.5cm] at (0,0){H\'enon Map} ;
\node [cell, fill=green!40, minimum height =2.cm, minimum width=2.5cm] at (3.5,0){H\'enon Map} ;
\node [cell, fill=green!40, minimum height =2.cm, minimum width=2.5cm] at (7,0){H\'enon Map} ;
\node [cell, fill=green!40, minimum height =2.cm, minimum width=2.5cm] at (10.5,0){H\'enon Map} ;

\end{scope}
          
\begin{scope}[xshift=17.8cm,yshift=2.4cm]
    \node [cell, fill=green!40, minimum height =4.7cm, minimum width=6cm] at (0,-.5){} ;
     \node [rectangle, dashed, draw,  thick, minimum height =2.6cm, minimum width=3.4cm] at (-0.5,-1.5){} ;
    \node [] at (0.,2.2) {\Large H\'enon Map};
     \node [] at (-1.8,-2.6) {FNN};
    
    \node [ circle,draw, line width = .75pt, minimum width=.6cm,inner sep=1pt,fill=red!50] (ibox4) at (0.6,-1.5) {$V(p)$};

    \node [operator] (nId) at (-0.5,1.5) {$-\cdot$};
    \node [operator] (mux2) at (-0.5,0.3) {$+\eta$};
    \node [operator] (mux3) at (2.4,-1.5) {$+$};
    \node [operator] (mux4) at (1.55,-1.5) {$\nabla$};
    
    \coordinate (x) at (-2.5,-0.75);
     \coordinate (y) at (1.5,1.5);
      \coordinate (z) at (2.7,.3);
       \coordinate (fnn) at (-1.8,-1.5);
       \node[] (cross) at (2.4,.3){};

    \node[ct, label={[mylabel]Input1}] (c) at (-4,1.5) {{\large$q$}};
    \node[ct, label={[mylabel]Input2}] (h) at (-4,-1.5) {{\large$p$}};
    

    \node[ct, label={[mylabel]Output1}] (c2) at (4,1.5) {\large$q^{[1]}$};
    \node[ct, label={[mylabel]Output2}] (h2) at (4,-1.5) {\large$p^{[1]}$};
    \foreach \name / \y in {1,...,4}
        \node[circle, minimum size=12pt,inner sep=0pt, fill=blue!50] (I-\name) at (-.5, .15-\y/1.5) {};
     
    \foreach \name / \y in {1,...,4}
         \draw [->, thin, draw=black!50] (I-\name)--(ibox4);
         
      \foreach \name / \y in {1,...,4}
          \draw [->, thin, draw=black!50] (fnn)-- (I-\name);
    \draw [ArrowC1] (c) -- (nId) -- (y);
    
    \draw [thick] (h) --(fnn);
    \draw [ArrowC1] (h -| x)++(-0.5,0) -| (x);

    \draw [-, ArrowC2] (x) |- (mux2);
    \draw [->, thick] (ibox4)--(mux4)-- (mux3);
    \draw [-, ArrowC1] (nId) -| (cross);
    \draw [->, thick] (cross)++(0.,-.15) -| (mux3);

    \draw [->, thick] (mux3)++(0.2,0) |- (h2);
    \draw [->, ArrowC2] (mux2)++(0.2,0) --(z) |- (c2);
    \end{scope}

\begin{scope}[xshift=1.4cm,yshift=6.5cm]
\node [backgroundred, minimum height =9.cm, minimum width=9.5cm] at (3.9,-2.7){} ;
  \node [] at (4,1.5) {\Large Symplectic Gyroceptron};
\node[ct, label={[mylabel]Input}] (c) at (0,-.5) {{\Large$(q, p)$}};
\node[ct, label={[mylabel]Output}] (o) at (6,-5) {{\Large$(\bar q, \bar p)$}};

\node[] (nn1p) at (3,-.8){};
\node[] (cross) at (1,-2){};
\node[] (center) at (6,-.5){};

 \node [cell, fill=red!40, minimum height =2.cm, minimum width=2.5cm] (nn1) at (3,-.5){\Large$\psi$} ;
  \node [cell, fill=red!40, minimum height =2.cm, minimum width=2.5cm] (nn2) at (3,-5){\Large$I_\epsilon$} ;
  \node [cell, fill=red!40, minimum height =2.cm, minimum width=3.5cm] (nn3) at (6.7,-.5){} ;
   \node [] at (8.,-1.2) {\Large$\Phi_\theta$};
   \node [] at (7.4,-.5) {\large$\theta$};
  
\draw[thick, ->] (center) arc (180:-155:.8);
\draw[dashed, ->] (6.4,-.5) arc (180:-155:.4);

    \draw [->, thick] (nn2)--(o);
     \draw [ArrowC1] (nn1p)++(-1.2,0) -| (cross);
      \draw [ArrowC1] (nn2)++(-1.2,0) -| (cross);
      \draw [->, thick] (1,-1)--(1,-3);
    \draw [->, thick] (c)--(nn1);
     \draw [->, thick] (nn1)--(nn3);
      \draw [->, thick] (5,-.8)--(4.2,-.8);
       \draw [->, thick] (6.1,-.8)--(5,-.8);
       \draw [->, thick] (5,-.5)--(6.05,-.5);
        \draw [dashed] (5,-.5)--(6.8,-.5);
        \draw [dashed] (6.1,-.8)--(6.8,-.5);
\end{scope}

 %
\end{tikzpicture}
}
\end{center}
\caption{Network diagrams. \  Left: Symplectic Gyroceptron. \  Right: H\'enon Network.
\label{fig:diagram}
\vspace{6mm} } 
\end{figure}
}

\hfill 

\section{Numerical Confirmation of the Existence of Adiabatic Invariants} \label{section: Numerical Confirmation of the Existence of Adiabatic Invariants}

\hfill 

In this section, we will confirm numerically that for any random set of weights and bias, the dynamical system generated by the symplectic gyroceptron
\begin{equation} \label{eq: Composition NN Adiabatic Section}  I_\varepsilon \ \circ \  \psi \ \circ \  \Phi_{\theta} \ \circ \  \psi^{-1} , \end{equation} introduced in Section~\ref{section: NN for NP Symplectic}, admits an adiabatic invariant. 

In our numerical experiments, we will take the circle action given by the clockwise rotation 
\begin{align} \label{eq: AdInv Circle Action}
	\mathcal{R}_{\theta}=\begin{pmatrix}
		\cos\theta & \sin\theta \\
		-\sin\theta & \cos\theta
	\end{pmatrix}.
\end{align}
The quantity $\mathfrak{I}_0(q,p) = \frac{1}{2} q^2 + \frac{1}{2} p^2$ is an invariant of the dynamics associated to the circle action~\eqref{eq: AdInv Circle Action}, and as a result \begin{equation} \label{eq: mu adiabatic} \mu \ = \  \mathfrak{I}_0 \ \circ \  \psi^{-1}
\end{equation}
is an invariant of the dynamics associated to the composition $ \psi \ \circ \  \Phi_{\theta} \ \circ \  \psi^{-1} ,$ and an adiabatic invariant of the dynamics associated to the symplectic gyroceptron~\eqref{eq: Composition NN Adiabatic Section}.

{
\newcommand{\figWidth}{8.5cm}
\newcommand{\trimfig}[2]{\trimw{#1}{#2}{.0}{.0}{.0}{.0}}
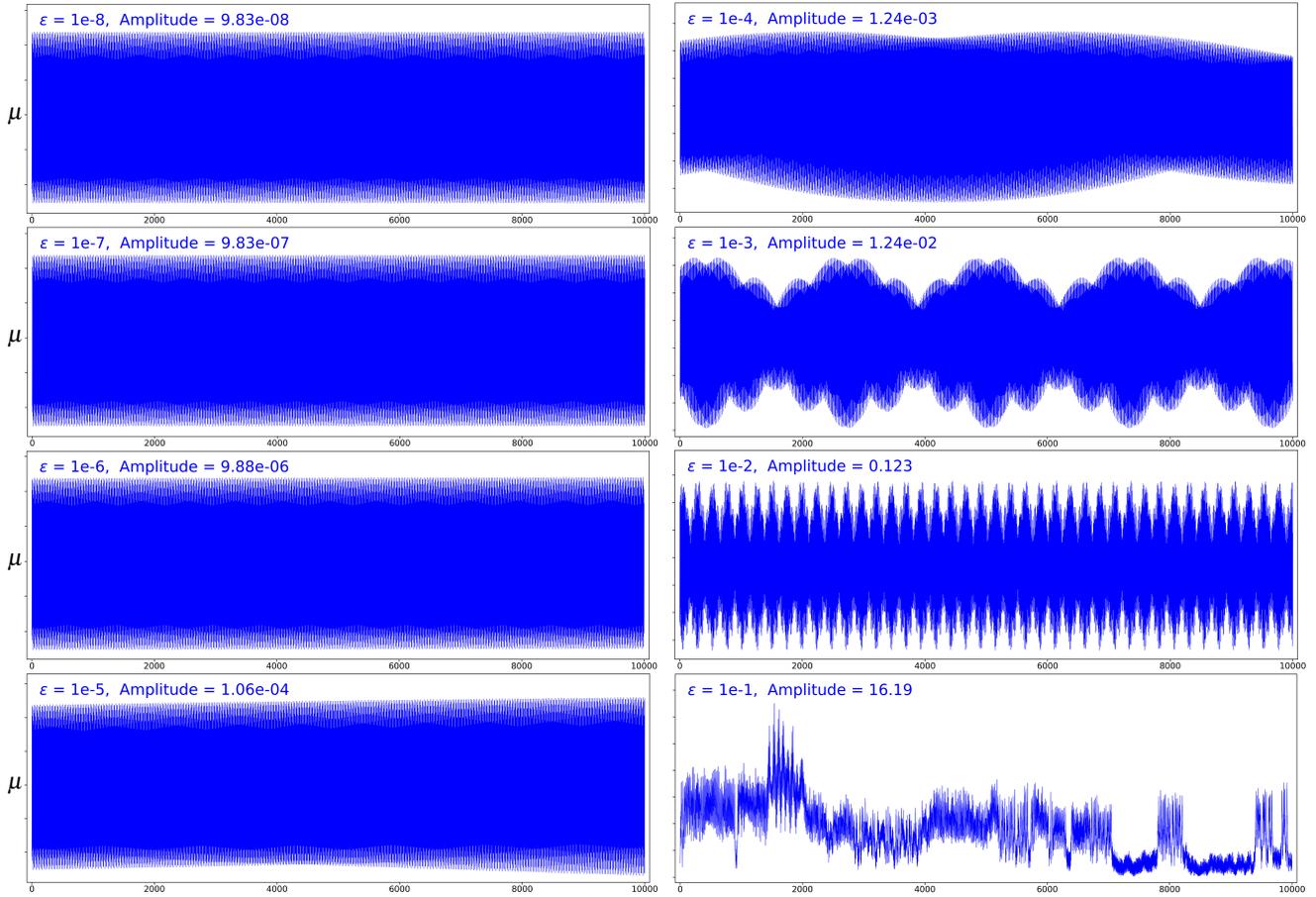
\begin{figure}[htb]
\begin{center}
\begin{tikzpicture}[scale=1]
  \useasboundingbox (0.,.7) rectangle (16,12.1);  
\draw(-1.,0) node[anchor=south west] {\trimfig{Profile_eps5}{\figWidth}};
\draw(7.7,0) node[anchor=south west] {\trimfig{Profile_eps1}{\figWidth}};
\node at (-.7,1.6) {\small$\mu$};
\begin{scope}[yshift=3cm]
\draw(-1.,0) node[anchor=south west] {\trimfig{Profile_eps6}{\figWidth}};
\draw(7.7,0) node[anchor=south west] {\trimfig{Profile_eps2}{\figWidth}};
\node at (-.7,1.6) {\small$\mu$};
\end{scope}
\begin{scope}[yshift=6cm]
\draw(-1.,0) node[anchor=south west] {\trimfig{Profile_eps7}{\figWidth}};
\draw(7.7,0) node[anchor=south west] {\trimfig{Profile_eps3}{\figWidth}};
\node at (-.7,1.6) {\small$\mu$};
\end{scope}
\begin{scope}[yshift=9cm]
\draw(-1.,0) node[anchor=south west] {\trimfig{Profile_eps8}{\figWidth}};
\draw(7.7,0) node[anchor=south west] {\trimfig{Profile_eps4}{\figWidth}};
\node at (-.7,1.6) {\small$\mu$};
\end{scope}
\end{tikzpicture}
\end{center}
\caption{ Conservation of the adiabatic invariant~\eqref{eq: mu adiabatic} over 10000 iterations for the map generated by the symplectic gyroceptron~\eqref{eq: Composition NN Adiabatic Section} as $\varepsilon$ is increased.\label{fig: AdInv Profiles} }
\end{figure}
}

Figure~\ref{fig: AdInv Profiles} displays the evolution of the adiabatic invariant~\eqref{eq: mu adiabatic} over 10000 iterations of the dynamical system generated by the symplectic gyroceptron~\eqref{eq: Composition NN Adiabatic Section}, for different values of $\varepsilon$. Here, $\psi$ is a H\'enonNet and $I_\varepsilon$ a near-identity H\'enonNet, both with 3 H\'enon layers, each of which has 8 neurons in its single-hidden-layer fully-connected neural networks layer potential. We can clearly see that the conservation of the adiabatic invariant gets significantly better as $\varepsilon$ gets closer to 0, going from chaotic oscillations of large amplitude when $\varepsilon = 0.1$ to very regular oscillations of minute amplitude when $\varepsilon = 10^{-8}$. 

We investigated further by obtaining the number of iterations needed for the adiabatic invariant~$\mu$ to deviate significantly from its original value~$\mu_0$ as $\varepsilon$ is varied. More precisely, given a value of $\varepsilon$, we search for the smallest integer $N(\varepsilon)$ such that
\begin{equation}
	|\mu_{N(\varepsilon)} - \mu_0 | \ > \  \rho  \max_{k= 0, ...,  K(\varepsilon) }{	|\mu_{k} - \mu_0 | }, \quad \text{ where } \ K(\varepsilon) = \lfloor 10 +  \varepsilon^{-1/4} \rfloor.
\end{equation}
In other words, we record the first iteration where the value of the adiabatic invariant $\mu$ deviates from its original value $\mu_0$ by more than some constant factor $\rho >1$ of the maximum deviations experienced in the first few $K(\varepsilon)$ iterations. The results are plotted in Figure~\ref{fig: Adiabatic} for $\rho = 1.1$. 



{
\newcommand{\figWidth}{4.6cm}
\newcommand{\trimfig}[2]{\trimh{#1}{#2}{.0}{.0}{.0}{.0}}
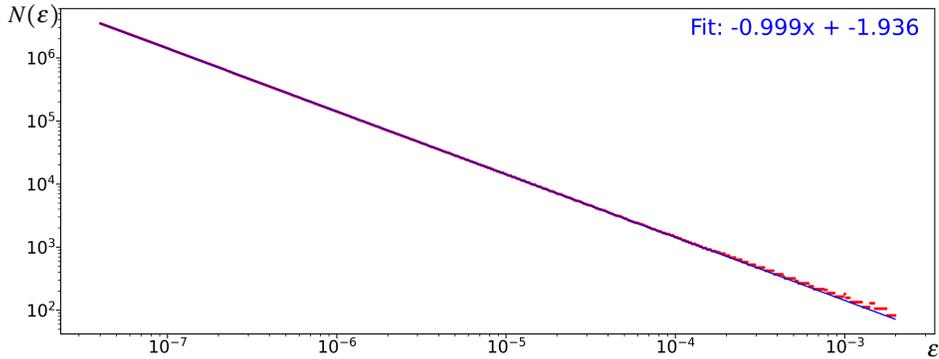
\begin{figure}[!ht]
\begin{center}
\begin{tikzpicture}[scale=1]
  \useasboundingbox (0.4,1.2) rectangle (13,5.3);  
  \draw(0,.5) node[anchor=south west,xshift=-4pt,yshift=+0pt] {\trimfig{Adiabatic}{\figWidth}};
 \node at (12.3,0.7) {\small$\epsilon$};
 \node at (.35,5.15) {\small$N(\epsilon)$};
\end{tikzpicture}
\end{center}
\caption{ $N(\varepsilon)$ as a function of $\varepsilon$ for $\rho = 1.1$ and a random set of weights. \label{fig: Adiabatic} }
\end{figure}
}

We can clearly see from Figure~\ref{fig: Adiabatic} that $N(\varepsilon)$, the number of iterations needed for the adiabatic invariant $\mu$ to deviate from its original value $\mu_0$ by more than $\rho = 1.1$ times the maximum deviations experienced in the first few iterations, increases sharply as $\varepsilon$ gets closer to 0. This is consistent with theoretical expectations. Note that using higher values of $\rho$ and smaller values of $\varepsilon$ would probably generate more interesting and meaningful results. Unfortunately, this is not computationally realizable since $N(\varepsilon)$ becomes very large when $\rho$ is increased beyond $1.2$. Even for larger values of~$\varepsilon$, computing a single point would take several days. \\

\section{Numerical Examples of Learning Surrogate Maps} \label{section: Numerical Example}

\subsection{Nonlinearly Coupled Oscillators}   \label{section: Nonlinearly Coupled Oscillators}

\hfill 

In this section, we use the symplectic gyroceptron architecture introduced in Section~\ref{section: NN for NP Symplectic} to learn a surrogate map for the nearly-periodic symplectic flow map associated to a nearly-periodic Hamiltonian system composed of two nonlinearly coupled oscillators, where one of them oscillates significantly faster than the other:
\begin{equation} \label{eq: Example of Interest Equations}	\begin{cases}
		\dot{q}_1 = p_1 \qquad & \dot{p}_1 = -q_1 - \varepsilon \partial_{q_1} U(q_1,q_2) \\ \dot{q}_2 = \varepsilon p_2 \qquad & \dot{p}_2 = -\varepsilon q_2 - \varepsilon \partial_{q_2} U(q_1,q_2)
\end{cases} \end{equation}
These equations of motion are the Hamilton's equations associated to the Hamiltonian
\begin{equation} 
	H_\varepsilon (q_1,q_2,p_1,p_2) = \frac{1}{2} (q_1^2 + p_1^2 ) + \frac{1}{2} \varepsilon (q_2^2 + p_2^2 )  + \varepsilon U(q_1,q_2).    
\end{equation}

The $\varepsilon = 0$ dynamics are decoupled, where the first oscillator, initialized at $\left(q_1(0),p_1(0)\right) = (\mathcal{q},\mathcal{p})$, follows a trajectory characterized by periodic clockwise circular rotation in phase space, while the second oscillator remains immobile:
\begin{equation}	
	q_1(t) = \mathcal{q} \cos{t} + \mathcal{p} \sin{t},   \qquad \qquad   p_1(t) = \mathcal{p} \cos{t} - \mathcal{q} \sin{t} .    
\end{equation}
Thus, this is a nearly-periodic Hamiltonian system on $\mathbb{R}^4$ with associated $\varepsilon = 0$ circle action given by the clockwise rotation
\begin{align*}
	\mathcal{R}_{\theta}=\begin{pmatrix}
		\cos\theta & 0 & \sin\theta & 0\\
		0 & 1 & 0 & 0\\ 
		-\sin\theta  & 0 & \cos\theta & 0 \\ 0 & 0 & 0 & 1
	\end{pmatrix}.
\end{align*}

We will use the nonlinear coupling potential $U(q_1,q_2) = q_1 q_2 \sin{(2q_1 + 2q_2)} $ in our numerical experiments since the resulting nearly-periodic Hamiltonian system displays complicated dynamics as the value of $\varepsilon$ is increased from 0. We have plotted in Figure~\ref{fig: Trajectories} a few trajectories of this dynamical system corresponding to different values of $\varepsilon$.   

\hfill

To learn a surrogate map for the nearly-periodic symplectic flow map associated to this nearly-periodic Hamiltonian system, we use the symplectic gyroceptron $  I_\varepsilon \ \circ \  \psi \ \circ \  \Phi_{\theta} \ \circ \  \psi^{-1} $ introduced in Section~\ref{section: NN for NP Symplectic}. In our numerical experiments, $\varepsilon = 0.01$, $\theta $ is a trainable parameter, $\psi$ is a H\'enonNet with 10 H\'enon layers each of which has 8 neurons in its single-hidden-layer fully-connected neural networks layer potential, and $I_\varepsilon$ is a near-identity H\'enonNet with 8 H\'enon layers each of which has 6 neurons in its single-hidden-layer fully-connected neural network layer potential.

The resulting symplectic gyroceptron of 549 trainable parameters was trained for a few thousands epochs on a dataset of 20,000 updates $(q_1,q_2,p_1,p_2) \mapsto(\tilde{q}_1,\tilde{q}_2,\tilde{p}_1,\tilde{p}_2) $ of the time-$0.05$ flow map associated to the nearly-periodic Hamiltonian system~\eqref{eq: Example of Interest Equations}. The training data was generated using the classical Runge--Kutta 4 integrator with very small time-steps, and the Mean Squared Error was used as the loss function in the training. Figure~\ref{fig: Learnt Dynamics} shows the dynamics predicted by the symplectic gyroceptron for seven different initial conditions with the same initial values of $(q_1,p_1)$ against the reference trajectories generated by the classical Runge--Kutta 4 integrator with very small time-steps. We only display the trajectories of the second oscillator since the motion of the first oscillator follows a simple nearly-circular curve.


We can see that the dynamics learnt by the symplectic gyroceptron match almost perfectly the reference trajectories and follow the level sets of the averaged Hamiltonian $\bar{H} = \frac{1}{2\pi}\int_0^{2\pi}\Phi_\theta^*H\,d\theta$, which is given by
\begin{align}
	\bar{H}(q_2,p_2) & = \frac{1}{2}(q_2^2 + p_2^2) + \frac{1}{2\pi} \int_0^{2\pi}{U(q_1(t) , q_2) dt}  \\ &= \frac{1}{2}(q_2^2 + p_2^2) +  \frac{q_2}{2\pi}\int_0^{2\pi}{ \left( \mathcal{q} \cos{t} +\mathcal{p} \sin{t} \right)  \sin{\left(2\left[ \mathcal{q} \cos{t} + \mathcal{p} \sin{t}\right]+ 2q_2 \right)} dt}   \\&=  \frac{1}{2}(q_2^2 + p_2^2) +  q_2 \cos{(2 q_2)}   \sqrt{ \mathcal{q}^2 + \mathcal{p}^2}   \ \mathcal{J}_1\left(2 \sqrt{ \mathcal{q}^2 + \mathcal{p}^2} \right) \label{eq: approximate averaged Hamiltonian} 
\end{align} 
where $\mathcal{J}_1(x)$ is the first order Bessel function of the first kind, up to an unimportant constant. Using Kruskal's theory of nearly-periodic systems, it is straightforward to show that this averaged Hamiltonian is the leading-order approximation of the Hamiltonian for the formal $U(1)$-reduction of the two-oscillator system.  \\

We also learned a surrogate map for the nearly-periodic symplectic time-$5$ flow map associated to the dynamical system~\eqref{eq: Example of Interest Equations}, using a symplectic gyroceptron where $\varepsilon = 0.01$, $\theta $ is a trainable parameter, and $\psi$ and $I_\varepsilon$ both have 10 H\'enon layers each of which has 8 neurons in its single-hidden-layer fully-connected neural network layer potential. This symplectic gyroceptron of 681 trainable parameters was trained for a few thousands epochs on a dataset of 60,000 updates $(q_1,q_2,p_1,p_2) \mapsto(\tilde{q}_1,\tilde{q}_2,\tilde{p}_1,\tilde{p}_2) $. For comparison, we also trained a H\'enonNet\cite{BurbyHenon} and a SympNet~\cite{Jin2020} of similar sizes and ran simulations from the same seven different initial conditions. The H\'enonNet used has 16 layers each of which has 10 neurons in its single-hidden-layer fully-connected neural network layer potential, for a total of 672 trainable parameters. The SympNet~\cite{Jin2020} used has 652 trainable parameters in a network structure of the form $ \mathcal{L}_n^{(k+1)}\ \circ \  (\mathcal{N}_{\text{up/low}}\ \circ \ \mathcal{L}_n^{(k)})\ \circ \ \dots \ \circ \ (\mathcal{N}_{\text{up/low}}\ \circ \ \mathcal{L}_n^{(1)})$, where each $\mathcal{L}_n^{(k)}$ is the composition of $n$ trainable linear symplectic layers, and $\mathcal{N}_{\text{up/low}}$ is a non-trainable symplectic activation map. \\


{
\newcommand{\figWidth}{8.5cm}
\newcommand{\trimfig}[2]{\trimw{#1}{#2}{.0}{.0}{.0}{.0}}
\begin{figure}[!htb]
\begin{center}
\begin{tikzpicture}[scale=1]
  \useasboundingbox (0.,.7) rectangle (16,17.5);  
\draw(-1.18,0) node[anchor=south west] {\trimfig{eps_0.11_Oscillator1}{\figWidth}};
\draw(7.59,0) node[anchor=south west] {\trimfig{eps_0.11_Oscillator2}{\figWidth}};
\node at (7.08,.6) {\small$(q_1, p_1)$};
\node at (15.83,.6) {\small$(q_2, p_2)$};
\begin{scope}[yshift=4.55cm]
\draw(-1.18,0) node[anchor=south west] {\trimfig{eps_0.07_Oscillator1}{\figWidth}};
\draw(7.59,0) node[anchor=south west] {\trimfig{eps_0.07_Oscillator2}{\figWidth}};
\node at (7.08,.6) {\small$(q_1, p_1)$};
\node at (15.83,.6) {\small$(q_2, p_2)$};
\end{scope}
\begin{scope}[yshift=9.1cm]
\draw(-1.18,0) node[anchor=south west] {\trimfig{eps_0.02_Oscillator1}{\figWidth}};
\draw(7.59,0) node[anchor=south west] {\trimfig{eps_0.02_Oscillator2}{\figWidth}};
\node at (7.06,.6) {\small$(q_1, p_1)$};
\node at (15.83,.6) {\small$(q_2, p_2)$};
\end{scope}
\begin{scope}[yshift=13.65cm]
\draw(-1.18,0) node[anchor=south west] {\trimfig{eps_0_Oscillator1}{\figWidth}};
\draw(7.59,0) node[anchor=south west] {\trimfig{eps_0_Oscillator2}{\figWidth}};
\node at (7.08,.6) {\small$(q_1, p_1)$};
\node at (15.83,.6) {\small$(q_2, p_2)$};
\end{scope}
\end{tikzpicture}
\end{center}
\vspace{2mm}
\caption{Sample trajectories in $(q_1,p_1)$ and $(q_2,p_2)$ phase spaces (left column: first oscillator, right column: second oscillator) for the nearly-periodic Hamiltonian system~\eqref{eq: Example of Interest Equations} as the value of the parameter $\varepsilon$ is increased.\label{fig: Trajectories}}
\end{figure}
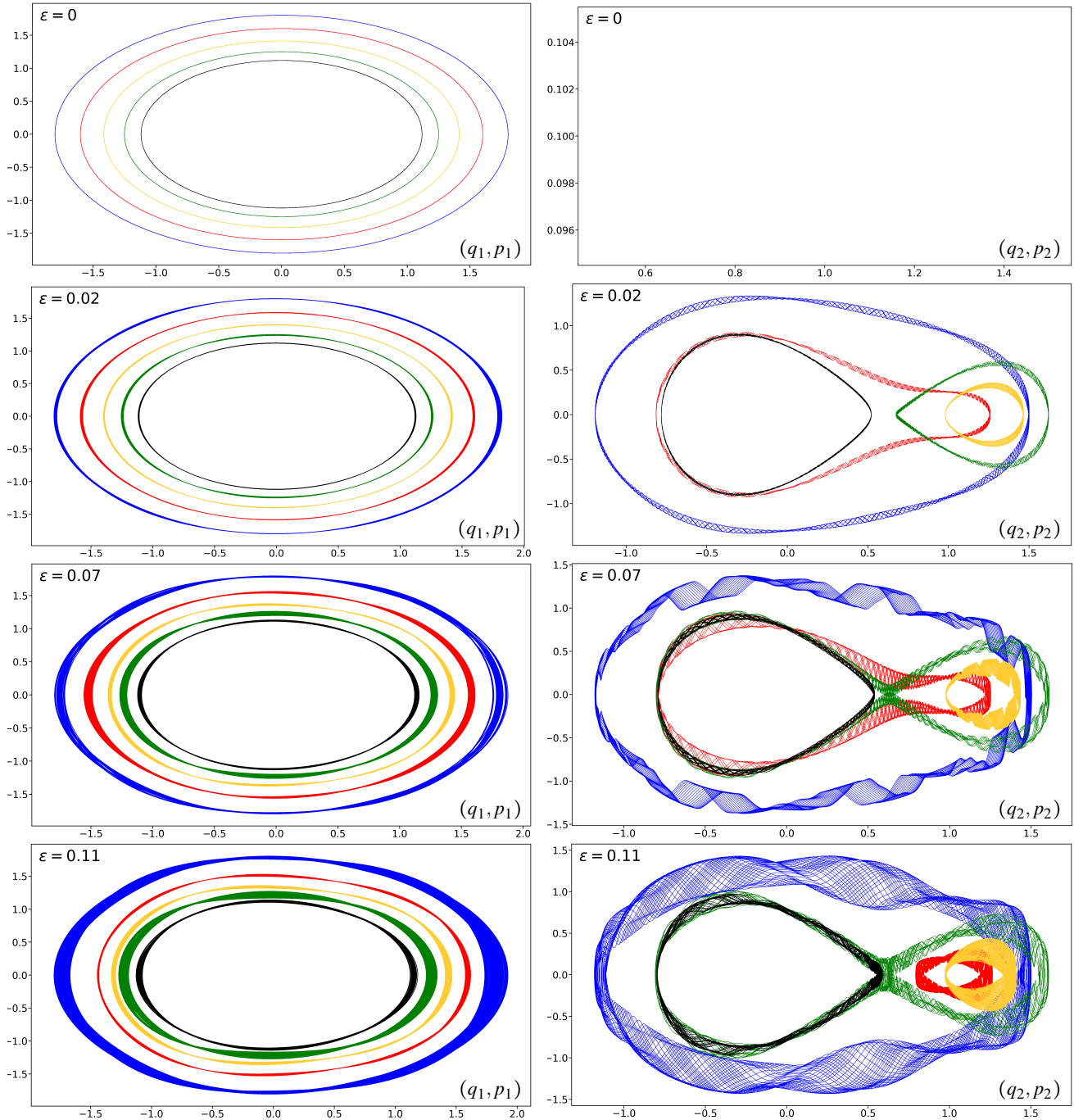
}

{
\newcommand{\figWidth}{8.6cm}
\newcommand{\figWidthb}{8.5cm}
\newcommand{\figWidthc}{9.5cm}
\newcommand{\trimfig}[2]{\trimw{#1}{#2}{.0}{.0}{.0}{.0}}
\begin{figure}[!ht]
\begin{center}
\begin{tikzpicture}[scale=1]
  \useasboundingbox (0.,.7) rectangle (16,9.7);  
\draw(-1.2,0) node[anchor=south west] {\trimfig{Reference_Oscillator2}{\figWidthb}};
\draw(7.7,0) node[anchor=south west] {\trimfig{NN_Model_Prediction_Oscillator2}{\figWidth}};
\node at (7.1,.6) {\small$(q_2, p_2)$};
\node at (16.1,.6) {\small$(q_2, p_2)$};
\draw(2.8,4.6) node[anchor=south west] {\trimfig{Level_Sets}{\figWidthc}};
\node at (12.,5.2) {\small$(q_2, p_2)$};
\end{tikzpicture}
\end{center}
\caption{Level sets of the averaged Hamiltonian~\eqref{eq: approximate averaged Hamiltonian}, and the symplectic gyroceptron predictions against the reference trajectories for the second oscillator in the nearly-periodic Hamiltonian system~\eqref{eq: Example of Interest Equations} with $\varepsilon = 0.01$ and a time-step of 0.05.\label{fig: Learnt Dynamics}}
\end{figure}
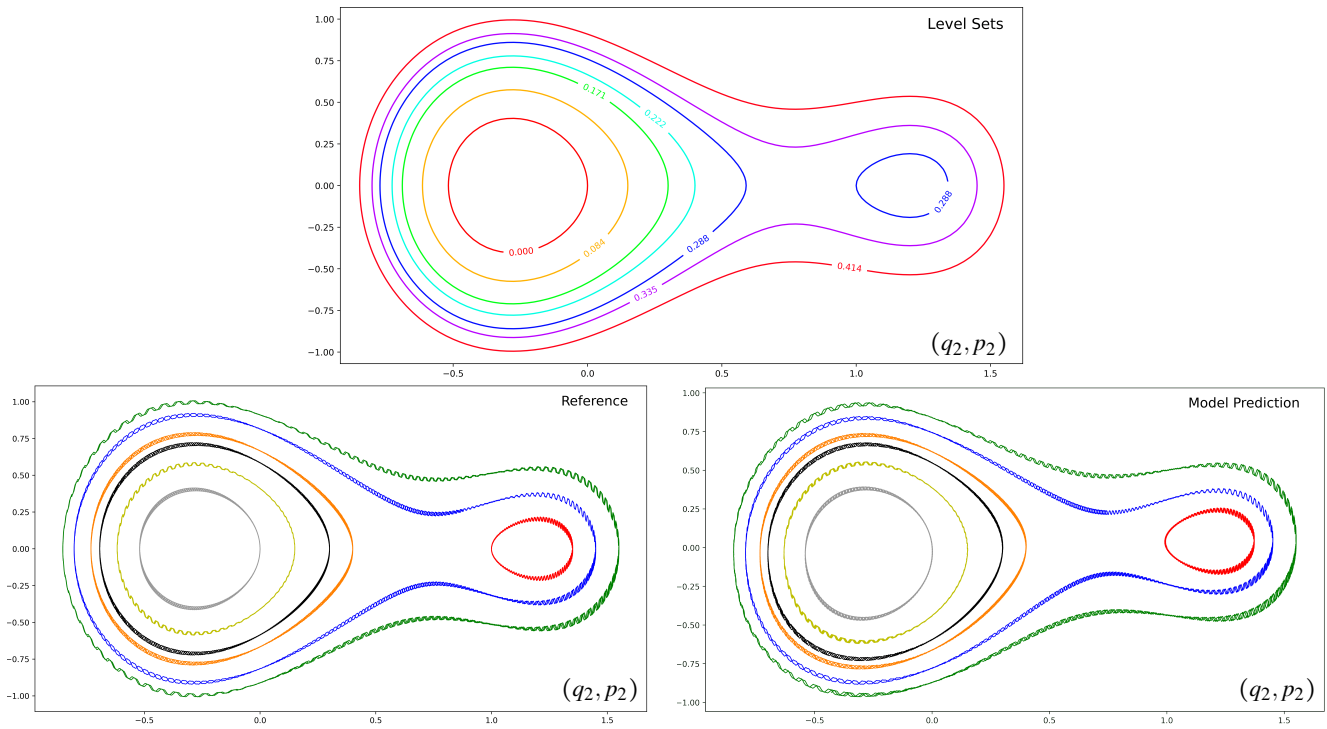
\vspace{-2mm}
}

{
\newcommand{\figWidth}{8.7cm}
\newcommand{\trimfig}[2]{\trimw{#1}{#2}{.0}{.0}{.0}{.0}}
\begin{figure}[!ht]
	\vspace{-2mm}
\begin{center}
\begin{tikzpicture}[scale=1]
  \useasboundingbox (0.,.7) rectangle (16,9.7);  
\draw(-1.2,0) node[anchor=south west] {\trimfig{LargeOscillator_SympNet}{\figWidth}};
\draw(7.65,0) node[anchor=south west] {\trimfig{LargeOscillator_HenonNet}{\figWidth}};
\node at (7.3,.6) {\small$(q_2, p_2)$};
\node at (16.2,.6) {\small$(q_2, p_2)$};
\begin{scope}[yshift=4.8cm]
\draw(-1.2,0) node[anchor=south west] {\trimfig{LargeOscillator_Reference}{\figWidth}};
\draw(7.65,0) node[anchor=south west] {\trimfig{LargeOscillator_SymplecticGyroceptron}{\figWidth}};
\node at (7.3,.6) {\small$(q_2, p_2)$};
\node at (16.2,.6) {\small$(q_2, p_2)$};
\end{scope}
\end{tikzpicture}
\end{center}\caption{Predictions from a Symplectic Gyroceptron, a SympNet, and a HenonNet, against the reference trajectories for the second oscillator in the nearly-periodic Hamiltonian system~\eqref{eq: Example of Interest Equations} with $\varepsilon = 0.01$ and the larger time-step of 5. \label{fig: Learnt Dynamics_BigStep}} \vspace{-3.5mm}
\end{figure}
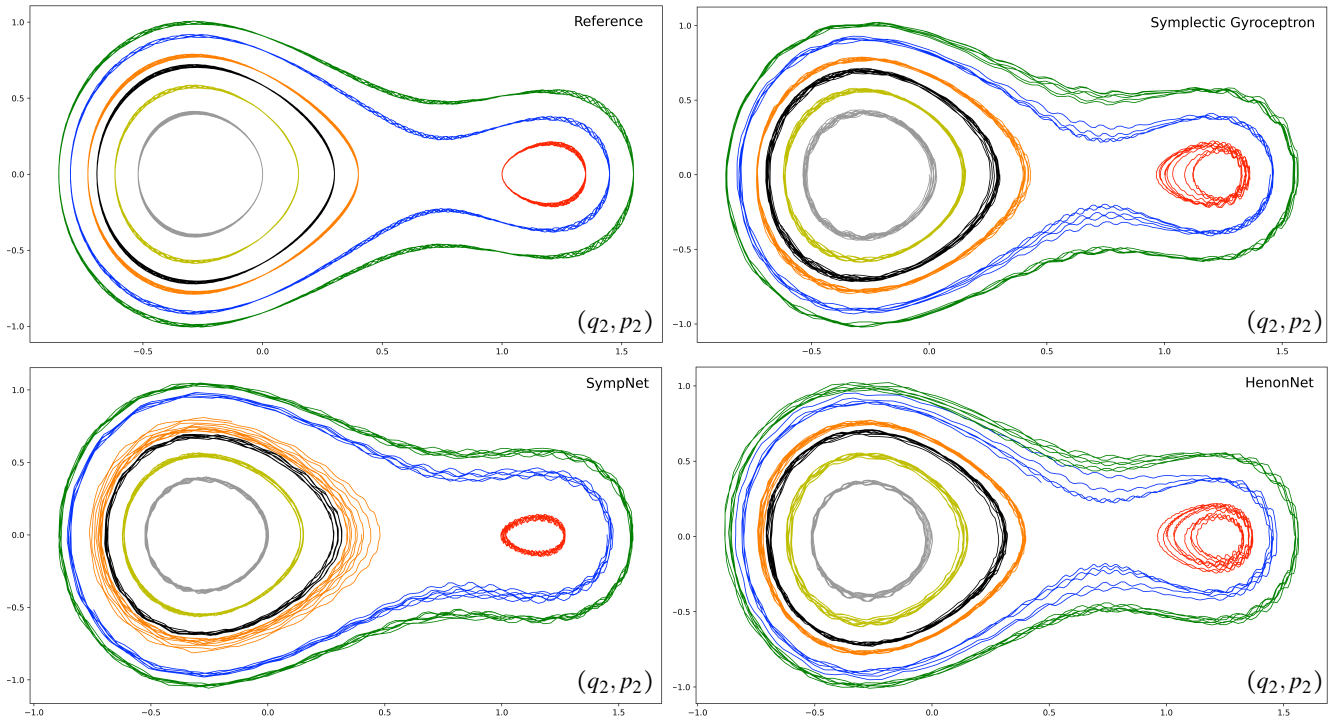
}

Figure~\ref{fig: Learnt Dynamics_BigStep} shows the dynamics predicted by the symplectic gyroceptron, the H\'enonNet, and the SympNet, for seven different initial conditions with the same initial values of $(q_1,p_1)$ against the reference trajectories generated by the Runge--Kutta 4 integrator (RK4) with small time-steps. As before, we only display the trajectories of the second oscillator. We can see that the dynamics predicted by the symplectic gyroceptron match the reference trajectories very well, although the predicted oscillations around the level sets of the averaged Hamiltonian are unsurprisingly larger than when learning the time-0.05 flow map. The crucial advantage that the symplectic gyroceptron offers over the other architectures considered, which only enforce the symplectic constraint, is provable existence of an adiabatic invariant. After training, the other architectures may empirically display preservation of an adiabatic invariant, but this cannot be proved rigorously from first principles. In contrast, the symplectic gyroceptron enjoys provable existence of an adiabatic invariant before, during, and after training. \\

Note that the symplectic gyroceptron generated the seven trajectories in 5 seconds, which is several orders of magnitude faster than RK4 with small time-steps which took 6,055 seconds. The H\'enonNet allowed to simulate the dynamics slightly faster, in 3 seconds, while the SympNet was much slower with a running time of 127 seconds, consistently with the observations made in the original H\'enonNet paper~\cite{BurbyHenon} which motivated choosing H\'enonNets over SympNets in the symplectic gyroceptrons. 

\newpage

\subsection{Charged Particle Interacting with its Self-Generated Electromagnetic Field} 
\label{sec: higher-dimensional example}

\vspace{1mm}

\subsubsection{Problem Formulation}

\hfill 

Next we test the ability of symplectic gyroceptrons to function as surrogates for higher-dimension nearly-periodic systems, and for systems where the limiting circle action is not precisely known. 

To formulate the ground-truth model first fix a positive integer $K$ and a sequence of single-variable functions $V_k:\mathbb{R}\rightarrow\mathbb{R}$, $k=1,\dots,K$. Consider the canonical Hamiltonian system on $ \mathbb{R}^2\times (\mathbb{R}^2)^K$ with coordinates $(q,p,Q_1,P_1,\dots,Q_K,P_K)$, defined by the Hamiltonian 
\begin{align}  \label{eq: Hamiltonian Higher Dim}
H_\epsilon = \frac{1}{2} \epsilon \left(p-\sum_{k=1}^K \sin(kq)\,Q_k\right)^2 + \frac{1}{2}\sum_{k=1}^Kk\left([P_k-V_k(Q_k)]^2+ Q_k^2\right).
\end{align}
The equations of motion are 
\begin{alignat*}{2}
& \dot{q}  = \partial_{p}H_\epsilon = \epsilon\,\left(p-\sum_{\ell =1}^K \sin(\ell q)\,Q_\ell \right) ,  \qquad  
 && \dot{p} = -\partial_{q}H_\epsilon =\epsilon\,\left(p-\sum_{\ell =1}^K \sin(\ell q)\,Q_\ell \right) \sum_{m=1}^K m \,\cos(m q)\,Q_m , \\
& \dot{Q}_k = \partial_{P_k}H_\epsilon = k\,(P_k - V_k(Q_k)) , 
&& \dot{P}_k  = -\partial_{Q_k}H_\epsilon =- k\,Q_k + k\,(P_k - V_k(Q_k))\,V_k^\prime(Q_k)+ \epsilon\,\left(p-\sum_{\ell =1}^K \sin(\ell q)\,Q_\ell \right)\ \sin(kq).
\end{alignat*}
These equations may be regarded as a simplified model of a charged particle $(q,p)$ interacting with its self-generated electromagnetic field $(Q_1,P_1,\dots,Q_K,P_K)$. We will describe the application of symplectic gyroceptrons to the development of a dynamical surrogate for this system when $\epsilon \ll 1$.

First, we verify that this Hamiltonian system is nearly-periodic, since this is the type of dynamical systems that symplectic gyroceptrons are designed to handle. So consider the limiting dynamics when $\epsilon = 0$. The equations of motion reduce to 
\begin{align}
\dot{q}  =0, \qquad 
\dot{p}  =0,  \qquad  
\dot{Q}_k = \partial_{P_k}H_\epsilon = k\,(P_k - V_k(Q_k)), \qquad 
\dot{P}_k  = -\partial_{Q_k}H_\epsilon =- k\,Q_k + k\,(P_k - V_k(Q_k))\,V_k^\prime(Q_k).
\end{align}
While these equations of motion may appear impenetrable at first glance, the symplectic transformation of variables given by $\Lambda_0^{-1}:(q,p,Q_1,P_1,\dots,Q_K,P_K)\mapsto (q,p,Q_1,\Pi_1,\dots,Q_K,\Pi_K)$ where $\Pi_k = P_k - V_k(Q_k)$ simplifies them dramatically into
\begin{align}
\dot{q}  =0, \qquad
\dot{p}  =0, \qquad
\dot{Q}_k = k\,\Pi_k, \qquad 
\dot{\Pi}_k  = - k\,Q_k,
\end{align}
which correspond to a family (indexed by $k$) of harmonic oscillators with angular frequencies $k$. The solution map in these nice variables is therefore $\Phi_t^0(q,p,Q_1,\Pi_1,\dots,Q_K,\Pi_K) =(q,p,Q_1(t),\Pi_1(t),\dots,Q_K(t),\Pi_K(t))  $, where
\begin{align} \label{eq: Rotation High Dim}
Q_k(t)  = \cos(k\,t)\,Q_k + \sin(k\,t)\,\Pi_k ,  \qquad 
\Pi_k(t)  = -\sin(k\,t)\,Q_k + \cos(k\,t)\,\Pi_k.
\end{align}
Note that $\Phi^0_t$ is periodic with minimal period $2\pi$. The solution map in terms of the original variables $(Q_k,P_k)$ is therefore $\Phi_t = \Lambda_0 \ \circ \  \Phi^0_\theta \ \circ  \ \Lambda_0^{-1}$. Since $\Phi_t$ is periodic in $t$ with minimal period $2\pi$ the ground-truth equations are Hamiltonian nearly-periodic. The leading-order adiabatic invariant is
\begin{align} \label{eq: Adiabatic High Dim}
\mu_0 \  = \ \frac{1}{2} \sum_{k=1}^K k\,(\Pi_k^2 + Q_k^2)  \ = \  \frac{1}{2} \sum_{k=1}^K k\,([P_k - V_k(Q_k)]^2 + Q_k^2).
\end{align}
Symplectic gyroceptrons are therefore well-suited to surrogate modeling for this system. \\

\subsubsection{Numerical Experiments}

\hfill 

Here, we learn the nearly-periodic Hamiltonian system~\eqref{eq: Hamiltonian Higher Dim} in the 6-dimensional case (i.e., $K=2$) with $  V_1(Q_1) = \frac{1}{2}  \sin(2Q_1) $ and $ V_2(Q_2) = \frac{1}{2} \exp{ (-5 Q_2^2)}$. In our symplectic gyroceptron $ I_\varepsilon \ \circ \  \psi \ \circ \  \Phi_{\theta} \ \circ \  \psi^{-1}$, the circle action $\Phi_{\theta}$ is taken to be the rotation in equation~\eqref{eq: Rotation High Dim} with $\theta $ treated as a trainable parameter, and the H\'enonNets $\psi$ and $I_\varepsilon$ both have 12 H\'enon layers each of which has 8 neurons in its single-hidden-layer fully-connected neural network layer potential. The resulting architecture of 1,033 trainable parameters was trained for a few thousands epochs on a dataset of 60,000 updates $(q,p,Q_1,P_1,Q_2,P_2) \mapsto(\tilde{q},\tilde{p},\tilde{Q}_1,\tilde{P}_2,\tilde{Q}_1,\tilde{P}_2) $.

To verify visually that we have learnt the dynamics successfully, we select initial conditions on the zero level set of the adiabatic invariant $\mu_0$. There, dynamics should remain on that slow manifold which is lower-dimensional and thus more easily portrayed. For the Hamiltonian system~\eqref{eq: Hamiltonian Higher Dim}, the slow manifold is the zero level set of $\mu_0 = 0$, which we can see from equation~\eqref{eq: Adiabatic High Dim}, is the set of points $(q,p,Q_1,P_1,Q_2,P_2)$ such that $Q_1 = Q_2 = 0$ and $P_1 = V_1(Q_1) = V_1(0), \ P_2 = V_2(Q_2) = V_2(0)$. 

\newpage 

\begin{figure}[!h]
	\vspace{1mm}
	\centering
	\includegraphics[width=0.495\textwidth]{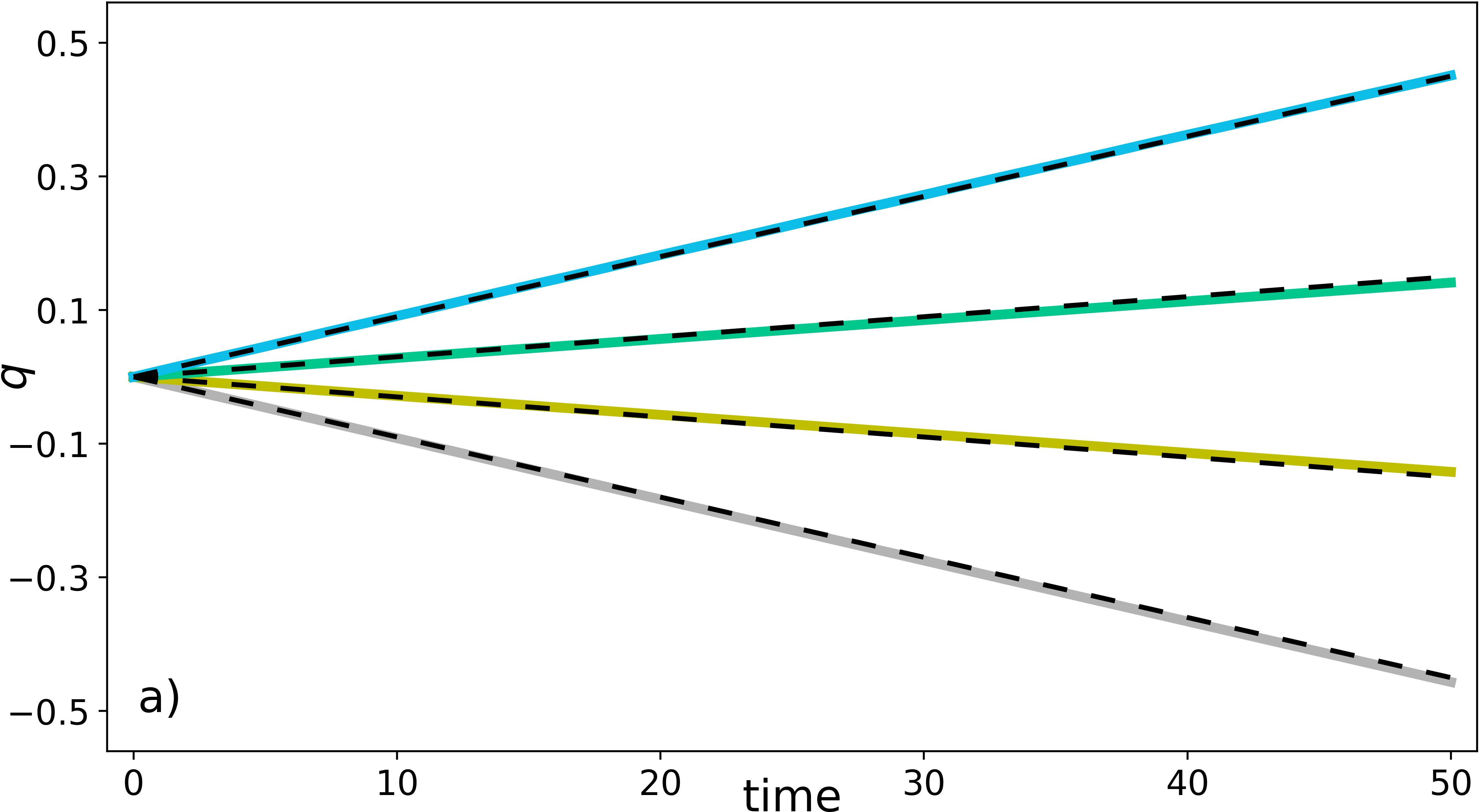} 
	\includegraphics[width=0.495\textwidth]{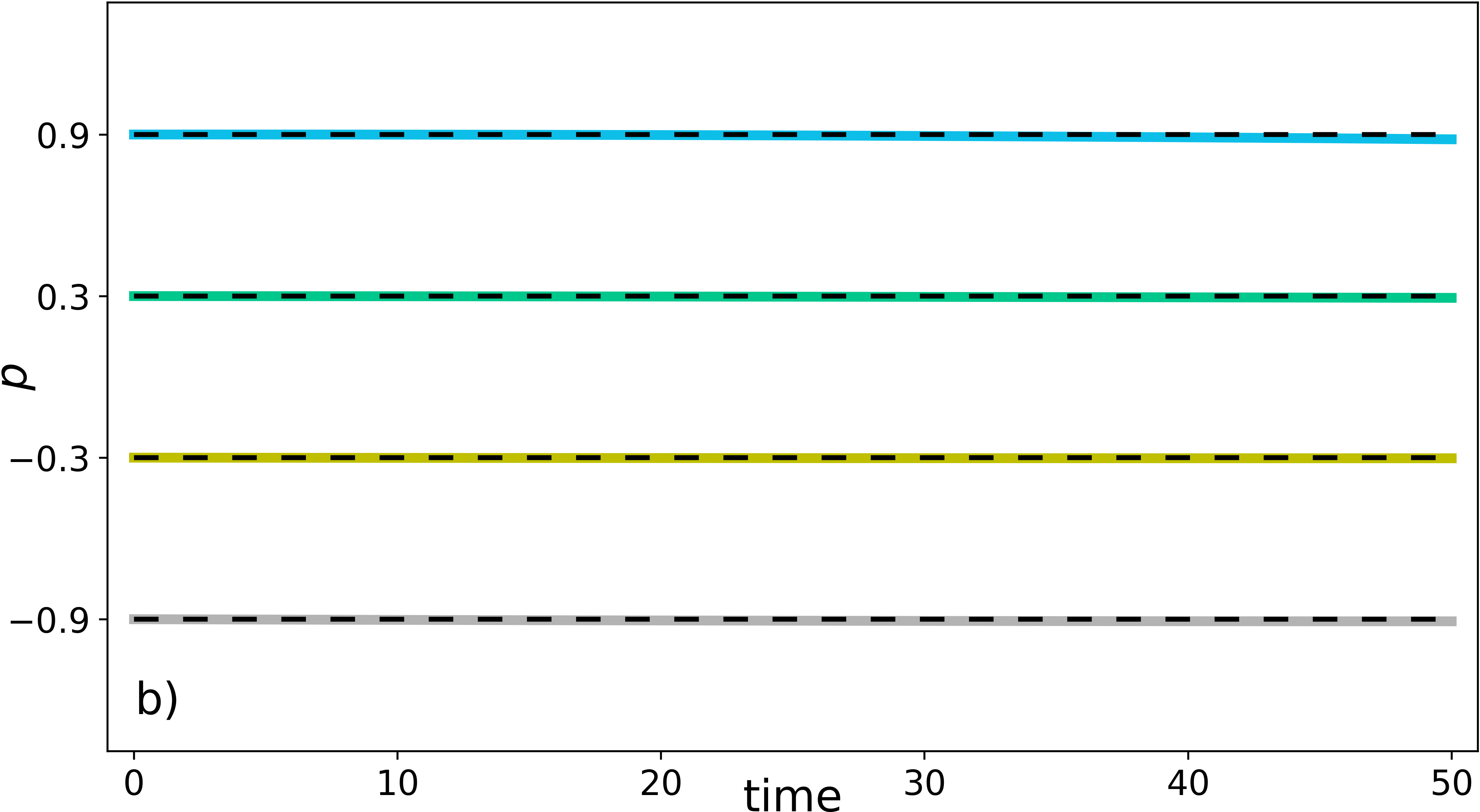}
	\vspace{2mm}
	
		\includegraphics[width=0.75\textwidth]{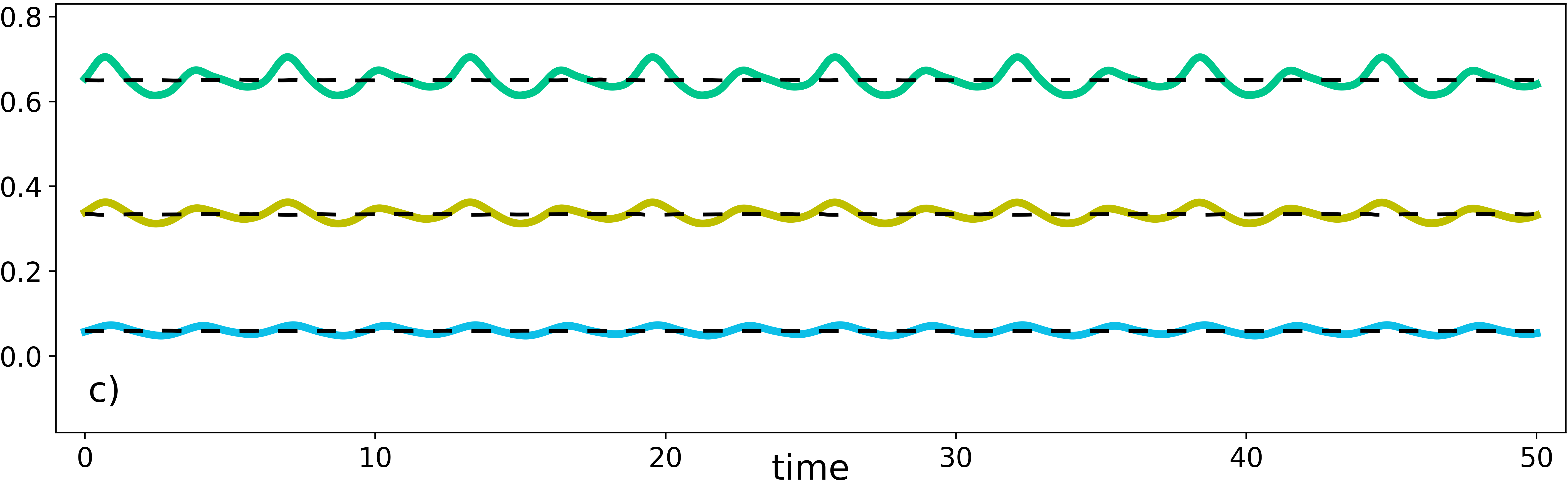} \vspace{-1mm}
	\caption{ \textbf{a) b)} Symplectic gyroceptron predictions (colors) against the true trajectories (dashed) with 4 different choices of initial conditions on the zero level set of the adiabatic invariant for the nearly-periodic Hamiltonian system~\eqref{eq: Hamiltonian Higher Dim} with $\varepsilon = 0.01$.  \textbf{c)} Evolution of $  \mathfrak{I}_0 \ \circ \  \psi^{-1} $  (colors) and $\mu^{learnt}_0$ (dashed lines) along trajectories generated by the symplectic gyroceptron with 3 different choices of initial conditions for the nearly-periodic Hamiltonian system~\eqref{eq: Hamiltonian Higher Dim} with $\varepsilon = 0.01$.     \label{fig: HD} }  
\end{figure}

On that slow manifold, the dynamics reduce to
\begin{align}
	 \dot{q}  = \epsilon  p ,   \qquad  
	 \dot{p} =0 \qquad \dot{Q}_1 = 0, \qquad \dot{Q}_2 = 0,  \qquad  \dot{P}_1  = \epsilon p \sin(q)   ,  \qquad  \dot{P}_2  = \epsilon p \sin(2q) ,
\end{align} 
where in particular the $(q,p)$ dynamics are now independent of $(Q_1, Q_2, P_1, P_2)$ and can easily be solved for explicitly, given some initial conditions $\left(q(0),p(0)\right) = (\mathcal{q},\mathcal{p})$:
\begin{equation}
	  q(t) =  \mathcal{q} + \epsilon \mathcal{p} t, \qquad  p(t) = \mathcal{p}.
\end{equation}
Figures~\ref{fig: HD}a)b) show that the trained symplectic gyroceptron generates predictions for the evolution of $q$ and $p$ which remain very close to the true trajectories on the slow manifold when the initial conditions are selected on the zero level set of $\mu_0$.

We also generate dynamics outside the zero level set of $\mu_0$ and verify that the quantity $ \mathfrak{I}_0 \ \circ \  \psi^{-1}$ matches the learnt adiabatic invariant $\mu^{learnt}_0 $ along the trajectories generated by the symplectic gyroceptron $  I_\varepsilon \ \circ \  \psi \ \circ \  \Phi_{\theta} \ \circ \  \psi^{-1} $, where
\begin{equation} \mu^{learnt}_0 (q,p,Q_1,P_1, Q_2,P_2) = \frac{1}{2} \sum_{k=1}^{K=2}k\,([P_k - V_k(Q_k)]^2 + Q_k^2) , \quad  \text{and}  \quad  \mathfrak{I}_0 (q,p,Q_1, \Pi_1, Q_2, \Pi_2) = \frac{1}{2} \sum_{k=1}^{K=2} k\,(\Pi_k^2 + Q_k^2)  . \end{equation} 
More precisely, we check whether $ \mathfrak{I}_0 \ \circ \  \psi^{-1} = \mu^{learnt}_0 $ with both quantities being approximately constant along trajectories generated by the symplectic gyroceptron, where 
	\begin{equation} 
	\mathfrak{I}_0 \ \circ \ \psi^{-1}(q,p,Q_1,P_1, Q_2,P_2)  = 	\frac{1}{2} \sum_{k=1}^{K=2} k\,(\tilde{\Pi}_k^2 + \tilde{Q}_k^2)  ,  \quad  \text{with } \ \ (\tilde{q}, \tilde{p} ,\tilde{Q_1}, \tilde{\Pi}_k ,\tilde{Q_2},  \tilde{\Pi}_2 )= \psi^{-1}(q,p,Q_1,P_1, Q_2,P_2). \end{equation}  
  From Figure~\ref{fig: HD}c), we see that along trajectories which are not started on the zero level set of $\mu_0$, the value of $ \mathfrak{I}_0 \ \circ \  \psi^{-1} $ remains close to the approximately constant quantity $ \mu^{learnt}_0 $, although $\mathfrak{I}_0 \ \circ \  \psi^{-1} $ displays small oscillations. Since $\mathfrak{I}_0 \ \circ \  \psi^{-1} $ is an adiabatic invariant for the network, these oscillations remain bounded in amplitude for very large time intervals. The amplitude can in principle be reduced by finding a more optimal set of weights for the network, but it can never be reduced to zero since the true adiabatic invariant is not exactly conserved (oscillations in $\mu_0$ are not visible at the scales displayed in the plot).

\section*{Discussion}

In this paper, we have successfully constructed novel structure-preserving neural network architectures, gyroceptrons and symplectic gyroceptrons, to learn nearly-periodic maps and nearly-periodic symplectic maps, respectively. 
By construction, these proposed architectures define nearly-periodic maps, and symplectic gyroceptrons also preserve symplecticity. Furthermore, it was confirmed experimentally that in the symplectic case, the maps generated by the proposed symplectic gyroceptrons admit discrete-time adiabatic invariants, regardless of the value of their parameters and weights. 

We also demonstrated that the proposed architectures can be effectively used in practice, by learning very precisely surrogate maps for the nearly-periodic symplectic flow maps associated to two different nearly-periodic Hamiltonian systems. Note that the hyperparameters in our architectures have not been optimized to maximize the quality of our training outcomes, and future applications of this architecture may benefit from further hyperparameter tuning. 

Symplectic gyroceptrons provide a promising class of architectures for surrogate modeling of non-dissipative dynamical systems that automatically steps over short timescales without introducing spurious instabilities, and could have potential future applications for the Klein--Gordon equation in the weakly-relativistic regime, for charged particles moving through a strong magnetic field, and for the rotating inviscid Euler equations in quasi-geostrophic scaling~\cite{Cotter_2004}. Symplectic gyroceptrons could also be used for structure-preserving simulation of non-canonical Hamiltonian systems on exact symplectic manifolds~\cite{BurbyLeok2021}, which have numerous applications across the physical sciences, for instance in modeling weakly-dissipative plasma systems~\cite{Morrison_1980,Morrison_MHD_1980,Morrison_fluid_1998,Burby_gvm_2015,Morrison_gen_beatification_2016,Morrison_neg_modes_1989,BurbyPOP2022}.  

The approach to symplectic gyroceptrons presented here targets surrogate modeling problems, where the dynamical system of interest is known but slow or expensive to simulate. In principle, symplectic gyroceptrons could also be used to discover dynamical models from observational data without detailed knowledge of the underlying dynamical system. However, in order to apply symplectic gyroceptrons effectively in this context data-mining methods must be developed for learning the topological conjugacy class of the limiting circle action. Given a topological classification of circle actions on the relevant state space (e.g. see \cite{Raymond_1968} for the case of a $3$-dimensional state space), a straightforward approach would be to test an ensemble of topologically-distinct circle actions for best results. A more nuanced approach would use the observed dynamics to estimate values for the classifying topological invariants of a circle action. This topological learning problem warrants further investigation. \\

\section*{Data Availability}

A simplified implementation of the Python codes used to generate some of the numerical results presented in this paper is published~\cite{Duruisseaux2023NPMapCode} and available at
	\begin{center}
		\url{https://github.com/vduruiss/SymplecticGyroceptron}
	\end{center} 

\hfill

\bibliography{Bibliography}

\def\cprime{$'$}
\begin{thebibliography}{10}
\urlstyle{rm}
\expandafter\ifx\csname url\endcsname\relax
  \def\url#1{\texttt{#1}}\fi
\expandafter\ifx\csname urlprefix\endcsname\relax\def\urlprefix{URL }\fi
\expandafter\ifx\csname doiprefix\endcsname\relax\def\doiprefix{DOI: }\fi
\providecommand{\bibinfo}[2]{#2}
\providecommand{\eprint}[2][]{\url{#2}}

\bibitem{karniadakis2021physics}
\bibinfo{author}{Karniadakis, G.~E.} \emph{et~al.}
\newblock \bibinfo{journal}{\bibinfo{title}{Physics-informed machine
  learning}}.
\newblock {\emph{\JournalTitle{Nature Reviews Physics}}}
  \textbf{\bibinfo{volume}{3}}, \bibinfo{pages}{422--440}
  (\bibinfo{year}{2021}).

\bibitem{BurbyHenon}
\bibinfo{author}{Burby, J.~W.}, \bibinfo{author}{Tang, Q.} \&
  \bibinfo{author}{Maulik, R.}
\newblock \bibinfo{journal}{\bibinfo{title}{Fast neural {P}oincar{\'{e}} maps
  for toroidal magnetic fields}}.
\newblock {\emph{\JournalTitle{Plasma Physics and Controlled Fusion}}}
  \textbf{\bibinfo{volume}{63}}, \bibinfo{pages}{024001},
  \doiprefix\url{10.1088/1361-6587/abcbaa} (\bibinfo{year}{2020}).

\bibitem{Jin2020}
\bibinfo{author}{Jin, P.}, \bibinfo{author}{Zhang, Z.}, \bibinfo{author}{Zhu,
  A.}, \bibinfo{author}{Tang, Y.} \& \bibinfo{author}{Karniadakis, G.~E.}
\newblock \bibinfo{journal}{\bibinfo{title}{{SympNets: Intrinsic
  structure-preserving symplectic networks for identifying Hamiltonian
  systems}}}.
\newblock {\emph{\JournalTitle{Neural Networks}}}
  \textbf{\bibinfo{volume}{132}}, \doiprefix\url{10.1016/j.neunet.2020.08.017}
  (\bibinfo{year}{2020}).

\bibitem{Willard2020}
\bibinfo{author}{Willard, J.~D.}, \bibinfo{author}{Jia, X.},
  \bibinfo{author}{Xu, S.}, \bibinfo{author}{Steinbach, M.~S.} \&
  \bibinfo{author}{Kumar, V.}
\newblock \bibinfo{title}{Integrating physics-based modeling with machine
  learning: A survey} (\bibinfo{year}{2020}).

\bibitem{Lei2020}
\bibinfo{author}{Lei, H.}, \bibinfo{author}{Wu, L.} \& \bibinfo{author}{Weinan,
  E.}
\newblock \bibinfo{journal}{\bibinfo{title}{Machine-learning-based
  non-{N}ewtonian fluid model with molecular fidelity}}.
\newblock {\emph{\JournalTitle{Physical Review E}}}
  \textbf{\bibinfo{volume}{102}}, \bibinfo{pages}{043309}
  (\bibinfo{year}{2020}).

\bibitem{Qin2020}
\bibinfo{author}{Qin, H.}
\newblock \bibinfo{journal}{\bibinfo{title}{Machine learning and serving of
  discrete field theories}}.
\newblock {\emph{\JournalTitle{Scientific Reports}}}
  \textbf{\bibinfo{volume}{10}}, \bibinfo{pages}{1--15} (\bibinfo{year}{2020}).

\bibitem{Cotter_2004}
\bibinfo{author}{Cotter, C.~J.} \& \bibinfo{author}{Reich, S.}
\newblock \bibinfo{journal}{\bibinfo{title}{Adiabatic invariance and
  applications: From molecular dynamics to numerical weather prediction}}.
\newblock {\emph{\JournalTitle{BIT Numer. Math.}}}
  \textbf{\bibinfo{volume}{44}}, \bibinfo{pages}{439} (\bibinfo{year}{2004}).

\bibitem{BurbyHi2021}
\bibinfo{author}{Burby, J.~W.} \& \bibinfo{author}{Hirvijoki, E.}
\newblock \bibinfo{journal}{\bibinfo{title}{Normal stability of slow manifolds
  in nearly periodic {H}amiltonian systems}}.
\newblock {\emph{\JournalTitle{Journal of Mathematical Physics}}}
  \textbf{\bibinfo{volume}{62}}, \bibinfo{pages}{093506},
  \doiprefix\url{10.1063/5.0054323} (\bibinfo{year}{2021}).

\bibitem{Kruskal1962}
\bibinfo{author}{Kruskal, M.}
\newblock \bibinfo{journal}{\bibinfo{title}{Asymptotic theory of {H}amiltonian
  and other systems with all solutions nearly periodic}}.
\newblock {\emph{\JournalTitle{Journal of Mathematical Physics}}}
  \textbf{\bibinfo{volume}{3}}, \bibinfo{pages}{806--828},
  \doiprefix\url{10.1063/1.1724285} (\bibinfo{year}{1962}).

\bibitem{BurbySquire_2020}
\bibinfo{author}{Burby, J.~W.} \& \bibinfo{author}{Squire, J.}
\newblock \bibinfo{journal}{\bibinfo{title}{General formulas for adiabatic
  invariants in nearly periodic hamiltonian systems}}.
\newblock {\emph{\JournalTitle{Journal of Plasma Physics}}}
  \textbf{\bibinfo{volume}{86}}, \bibinfo{pages}{835860601}
  (\bibinfo{year}{2020}).

\bibitem{BurbyLeok2021}
\bibinfo{author}{Burby, J.~W.}, \bibinfo{author}{Hirvijoki, E.} \&
  \bibinfo{author}{Leok, M.}
\newblock \bibinfo{title}{Nearly-periodic maps and geometric integration of
  noncanonical {H}amiltonian systems} (\bibinfo{year}{2021}).

\bibitem{Poincare1899}
\bibinfo{author}{Poincar\'e, H.}
\newblock \emph{\bibinfo{title}{{Les m\'ethodes nouvelles de la m\'ecanique
  c\'eleste, Volume 3}}} (\bibinfo{publisher}{Gauthier-Villars},
  \bibinfo{address}{Paris}, \bibinfo{year}{1899}).

\bibitem{HaLuWa2006}
\bibinfo{author}{Hairer, E.}, \bibinfo{author}{Lubich, C.} \&
  \bibinfo{author}{Wanner, G.}
\newblock \emph{\bibinfo{title}{Geometric {N}umerical {I}ntegration}},
  vol.~\bibinfo{volume}{31} of \emph{\bibinfo{series}{Springer Series in
  Computational Mathematics}} (\bibinfo{publisher}{Springer-Verlag},
  \bibinfo{address}{Berlin}, \bibinfo{year}{2006}), \bibinfo{edition}{second}
  edn.

\bibitem{IserlesWhyGNI}
\bibinfo{author}{Iserles, A.} \& \bibinfo{author}{Quispel, G.}
\newblock \bibinfo{title}{Why geometric numerical integration?}
  (\bibinfo{year}{2018}).

\bibitem{Blanes2017}
\bibinfo{author}{Blanes, S.} \& \bibinfo{author}{Casas, F.}
\newblock \emph{\bibinfo{title}{A Concise Introduction to Geometric Numerical
  Integration}} (\bibinfo{year}{2017}).

\bibitem{LeRe2005}
\bibinfo{author}{Leimkuhler, B.} \& \bibinfo{author}{Reich, S.}
\newblock \emph{\bibinfo{title}{Simulating {H}amiltonian Dynamics}},
  vol.~\bibinfo{volume}{14} of \emph{\bibinfo{series}{Cambridge Monographs on
  Applied and Computational Mathematics}} (\bibinfo{publisher}{Cambridge
  University Press}, \bibinfo{address}{Cambridge}, \bibinfo{year}{2004}).

\bibitem{Holm2009}
\bibinfo{author}{Holm, D.}, \bibinfo{author}{Schmah, T.} \&
  \bibinfo{author}{Stoica, C.}
\newblock \emph{\bibinfo{title}{Geometric Mechanics and Symmetry: From Finite
  to Infinite Dimensions}}.
\newblock Oxford Texts in Applied and Engineering Mathematics
  (\bibinfo{publisher}{OUP Oxford}, \bibinfo{year}{2009}).

\bibitem{Chen2020}
\bibinfo{author}{{Chen}, Z.}, \bibinfo{author}{{Zhang}, J.},
  \bibinfo{author}{{Arjovsky}, M.} \& \bibinfo{author}{{Bottou}, L.}
\newblock \bibinfo{title}{{Symplectic Recurrent Neural Networks}}.
\newblock In \emph{\bibinfo{booktitle}{International Conference on Learning
  Representations}} (\bibinfo{year}{2020}).

\bibitem{Chen2021neural}
\bibinfo{author}{Chen, Y.}, \bibinfo{author}{Matsubara, T.} \&
  \bibinfo{author}{Yaguchi, T.}
\newblock \bibinfo{title}{Neural symplectic form: learning {H}amiltonian
  equations on general coordinate systems}.
\newblock In \emph{\bibinfo{booktitle}{Advances in Neural Information
  Processing Systems}} (\bibinfo{year}{2021}).

\bibitem{Cranmer2020}
\bibinfo{author}{Cranmer, M.} \emph{et~al.}
\newblock \bibinfo{journal}{\bibinfo{title}{Lagrangian neural networks}}.
\newblock {\emph{\JournalTitle{ICLR Workshop on Integration of Deep Neural
  Models and Differential Equations}}}  (\bibinfo{year}{2020}).

\bibitem{Greydanus2019}
\bibinfo{author}{Greydanus, S.~l.}, \bibinfo{author}{Dzamba, M.} \&
  \bibinfo{author}{Yosinski, J.}
\newblock \bibinfo{title}{Hamiltonian neural networks}.
\newblock In \emph{\bibinfo{booktitle}{Advances in Neural Information
  Processing Systems}}, vol.~\bibinfo{volume}{32} (\bibinfo{year}{2019}).

\bibitem{Lutter2018}
\bibinfo{author}{Lutter, M.}, \bibinfo{author}{Ritter, C.} \&
  \bibinfo{author}{Peters, J.}
\newblock \bibinfo{title}{Deep {L}agrangian networks: Using physics as model
  prior for deep learning}.
\newblock In \emph{\bibinfo{booktitle}{International Conference on Learning
  Representations}} (\bibinfo{year}{2019}).

\bibitem{Zhong2020}
\bibinfo{author}{Zhong, Y.~D.}, \bibinfo{author}{Dey, B.} \&
  \bibinfo{author}{Chakraborty, A.}
\newblock \bibinfo{title}{{Symplectic ODE-Net: Learning Hamiltonian} dynamics
  with control}.
\newblock In \emph{\bibinfo{booktitle}{International Conference on Learning
  Representations}} (\bibinfo{year}{2020}).

\bibitem{zhong2020dissipative}
\bibinfo{author}{Zhong, Y.~D.}, \bibinfo{author}{Dey, B.} \&
  \bibinfo{author}{Chakraborty, A.}
\newblock \bibinfo{title}{Dissipative {S}ym{ODEN}: Encoding {H}amiltonian
  dynamics with dissipation and control into deep learning}.
\newblock In \emph{\bibinfo{booktitle}{ICLR 2020 Work. on Integration of Deep
  Neural Models and Differential Equations}} (\bibinfo{year}{2020}).

\bibitem{Zhong2021}
\bibinfo{author}{Zhong, Y.~D.}, \bibinfo{author}{Dey, B.} \&
  \bibinfo{author}{Chakraborty, A.}
\newblock \bibinfo{title}{Benchmarking energy-conserving neural networks for
  learning dynamics from data}.
\newblock In \emph{\bibinfo{booktitle}{Learning for Dynamics and Control}},
  vol. \bibinfo{volume}{144}, \bibinfo{pages}{1218--1229}
  (\bibinfo{year}{2021}).

\bibitem{Saemundsson2020}
\bibinfo{author}{S{\ae}mundsson, S.}, \bibinfo{author}{Terenin, A.},
  \bibinfo{author}{Hofmann, K.} \& \bibinfo{author}{Deisenroth, M.~P.}
\newblock \bibinfo{title}{Variational integrator networks for physically
  structured embeddings}.
\newblock In \emph{\bibinfo{booktitle}{AISTATS}} (\bibinfo{year}{2020}).

\bibitem{Havens2021}
\bibinfo{author}{Havens, A.} \& \bibinfo{author}{Chowdhary, G.}
\newblock \bibinfo{title}{Forced variational integrator networks for prediction
  and control of mechanical systems} (\bibinfo{year}{2021}).

\bibitem{LieFVINsExtended}
\bibinfo{author}{Duruisseaux, V.}, \bibinfo{author}{Duong, T.},
  \bibinfo{author}{Leok, M.} \& \bibinfo{author}{Atanasov, N.}
\newblock \bibinfo{title}{Lie group forced variational integrator networks for
  learning and control of robot systems} (\bibinfo{year}{2022}).

\bibitem{Santos2022}
\bibinfo{author}{Santos, S.}, \bibinfo{author}{Ekal, M.} \&
  \bibinfo{author}{Ventura, R.}
\newblock \bibinfo{title}{Symplectic momentum neural networks - using discrete
  variational mechanics as a prior in deep learning}.
\newblock In \emph{\bibinfo{booktitle}{Proceedings of The 4th Annual Learning
  for Dynamics and Control Conference}}, vol. \bibinfo{volume}{168} of
  \emph{\bibinfo{series}{Proceedings of Machine Learning Research}},
  \bibinfo{pages}{584--595} (\bibinfo{year}{2022}).

\bibitem{Valperga2022}
\bibinfo{author}{Valperga, R.}, \bibinfo{author}{Webster, K.},
  \bibinfo{author}{Turaev, D.}, \bibinfo{author}{Klein, V.} \&
  \bibinfo{author}{Lamb, J.}
\newblock \bibinfo{title}{Learning reversible symplectic dynamics}.
\newblock In \emph{\bibinfo{booktitle}{Proceedings of The 4th Annual Learning
  for Dynamics and Control Conference}}, vol. \bibinfo{volume}{168} of
  \emph{\bibinfo{series}{Proceedings of Machine Learning Research}},
  \bibinfo{pages}{906--916} (\bibinfo{year}{2022}).

\bibitem{Bertalan2019}
\bibinfo{author}{Bertalan, T.}, \bibinfo{author}{Dietrich, F.},
  \bibinfo{author}{Mezi{\'c}, I.} \& \bibinfo{author}{Kevrekidis, I.~G.}
\newblock \bibinfo{journal}{\bibinfo{title}{On learning {H}amiltonian systems
  from data}}.
\newblock {\emph{\JournalTitle{Chaos: An Interdisciplinary Journal of Nonlinear
  Science}}} \textbf{\bibinfo{volume}{29}}, \bibinfo{pages}{121107},
  \doiprefix\url{10.1063/1.5128231} (\bibinfo{year}{2019}).

\bibitem{Rath2021}
\bibinfo{author}{Rath, K.}, \bibinfo{author}{Albert, C.~G.},
  \bibinfo{author}{Bischl, B.} \& \bibinfo{author}{von Toussaint, U.}
\newblock \bibinfo{journal}{\bibinfo{title}{Symplectic gaussian process
  regression of maps in hamiltonian systems}}.
\newblock {\emph{\JournalTitle{Chaos: An Interdisciplinary Journal of Nonlinear
  Science}}} \textbf{\bibinfo{volume}{31}}, \bibinfo{pages}{053121},
  \doiprefix\url{10.1063/5.0048129} (\bibinfo{year}{2021}).

\bibitem{Offen2022}
\bibinfo{author}{Offen, C.} \& \bibinfo{author}{Ober-Bl{\"o}baum, S.}
\newblock \bibinfo{journal}{\bibinfo{title}{Symplectic integration of learned
  {H}amiltonian systems}}.
\newblock {\emph{\JournalTitle{Chaos: An Interdisciplinary Journal of Nonlinear
  Science}}} \textbf{\bibinfo{volume}{32}}, \bibinfo{pages}{013122},
  \doiprefix\url{10.1063/5.0065913} (\bibinfo{year}{2022}).

\bibitem{Marco2021}
\bibinfo{author}{Marco, D.} \& \bibinfo{author}{M{\'e}hats, F.}
\newblock \bibinfo{title}{Symplectic learning for {H}amiltonian neural
  networks} (\bibinfo{year}{2021}).

\bibitem{Mathiesen2022}
\bibinfo{author}{Mathiesen, F.~B.}, \bibinfo{author}{Yang, B.} \&
  \bibinfo{author}{Hu, J.}
\newblock \bibinfo{journal}{\bibinfo{title}{Hyperverlet: A symplectic
  hypersolver for {H}amiltonian systems}}.
\newblock {\emph{\JournalTitle{Proceedings of the AAAI Conference on Artificial
  Intelligence}}} \textbf{\bibinfo{volume}{36}}, \bibinfo{pages}{4575--4582},
  \doiprefix\url{10.1609/aaai.v36i4.20381} (\bibinfo{year}{2022}).

\bibitem{Morrison_1980}
\bibinfo{journal}{\bibinfo{author}{Morrison, P.~J.}}
\newblock {\emph{\JournalTitle{Phys. Lett.}}} \textbf{\bibinfo{volume}{80A}},
  \bibinfo{pages}{383} (\bibinfo{year}{1980}).

\bibitem{Morrison_MHD_1980}
\bibinfo{journal}{\bibinfo{author}{Morrison, P.~J.} \& \bibinfo{author}{Greene,
  J.~M.}}
\newblock {\emph{\JournalTitle{Phys. Rev. Lett.}}}
  \textbf{\bibinfo{volume}{45}}, \bibinfo{pages}{790} (\bibinfo{year}{1980}).

\bibitem{Morrison_fluid_1998}
\bibinfo{author}{Morrison, P.~J.}
\newblock \bibinfo{journal}{\bibinfo{title}{Nonlinear stability of fluid and
  plasma equilibria}}.
\newblock {\emph{\JournalTitle{Rev. Mod. Phys.}}}
  \textbf{\bibinfo{volume}{70}}, \bibinfo{pages}{467},
  \doiprefix\url{http://dx.doi.org/10.1103/RevModPhys.70.467}
  (\bibinfo{year}{1998}).

\bibitem{Burby_gvm_2015}
\bibinfo{author}{Burby, J.~W.}, \bibinfo{author}{Brizard, A.~J.},
  \bibinfo{author}{Morrison, P.~J.} \& \bibinfo{author}{Qin, H.}
\newblock \bibinfo{journal}{\bibinfo{title}{Hamiltonian gyrokinetic
  vlasov-maxwell system}}.
\newblock {\emph{\JournalTitle{Phys. Lett. A}}} \textbf{\bibinfo{volume}{379}},
  \bibinfo{pages}{2073}, \doiprefix\url{doi:10.1016/j.physleta.2015.06.051}
  (\bibinfo{year}{2015}).

\bibitem{Morrison_gen_beatification_2016}
\bibinfo{author}{Morrison, P.~J.} \& \bibinfo{author}{Vanneste, J.}
\newblock \bibinfo{journal}{\bibinfo{title}{Weakly nonlinear dynamics in
  noncanonical hamiltonian systems with applications to fluids and plasmas}}.
\newblock {\emph{\JournalTitle{Ann. Phys.}}} \textbf{\bibinfo{volume}{368}},
  \bibinfo{pages}{117},
  \doiprefix\url{https://doi.org/10.1016/j.aop.2016.02.003}
  (\bibinfo{year}{2016}).

\bibitem{Morrison_neg_modes_1989}
\bibinfo{author}{{Morrison}, P.~J.} \& \bibinfo{author}{{Kotschenreuther}, M.}
\newblock \bibinfo{title}{{The free energy principle, negative energy modes,
  and stability}}.
\newblock \bibinfo{pages}{9--22} (\bibinfo{year}{1989}).

\bibitem{BurbyPOP2022}
\bibinfo{author}{Burby, J.~W.}
\newblock \bibinfo{journal}{\bibinfo{title}{Slow manifold reduction as a
  systematic tool for revealing the geometry of phase space}}.
\newblock {\emph{\JournalTitle{Phys. Plasmas}}} \textbf{\bibinfo{volume}{29}},
  \bibinfo{pages}{042102}, \doiprefix\url{10.1063/5.0084543}
  (\bibinfo{year}{2022}).

\bibitem{Hernandez2021}
\bibinfo{author}{Hernandez, Q.}, \bibinfo{author}{Badías, A.},
  \bibinfo{author}{González, D.}, \bibinfo{author}{Chinesta, F.} \&
  \bibinfo{author}{Cueto, E.}
\newblock \bibinfo{journal}{\bibinfo{title}{Deep learning of
  thermodynamics-aware reduced-order models from data}}.
\newblock {\emph{\JournalTitle{Computer Methods in Applied Mechanics and
  Engineering}}} \textbf{\bibinfo{volume}{379}}, \bibinfo{pages}{113763},
  \doiprefix\url{https://doi.org/10.1016/j.cma.2021.113763}
  (\bibinfo{year}{2021}).

\bibitem{Hernandez2023}
\bibinfo{author}{Hernández, Q.}, \bibinfo{author}{Badias, A.},
  \bibinfo{author}{Chinesta, F.} \& \bibinfo{author}{Cueto, E.}
\newblock \bibinfo{journal}{\bibinfo{title}{Port-metriplectic neural networks:
  thermodynamics-informed machine learning of complex physical systems}}.
\newblock {\emph{\JournalTitle{Computational Mechanics}}}
  \bibinfo{pages}{1--9}, \doiprefix\url{10.1007/s00466-023-02296-w}
  (\bibinfo{year}{2023}).

\bibitem{Huang2022}
\bibinfo{author}{Huang, S.}, \bibinfo{author}{He, Z.}, \bibinfo{author}{Chem,
  B.} \& \bibinfo{author}{Reina, C.}
\newblock \bibinfo{journal}{\bibinfo{title}{{Variational Onsager Neural
  Networks (VONNs)}: A thermodynamics-based variational learning strategy for
  non-equilibrium {PDE}s}}.
\newblock {\emph{\JournalTitle{Journal of the Mechanics and Physics of
  Solids}}} \textbf{\bibinfo{volume}{163}}, \bibinfo{pages}{104856},
  \doiprefix\url{https://doi.org/10.1016/j.jmps.2022.104856}
  (\bibinfo{year}{2022}).

\bibitem{unicornn}
\bibinfo{author}{Rusch, T.~K.} \& \bibinfo{author}{Mishra, S.}
\newblock \bibinfo{title}{Unicornn: A recurrent model for learning very long
  time dependencies}.
\newblock In \emph{\bibinfo{booktitle}{Proceedings of the 38th Int. Conf. on
  Machine Learning}}, vol. \bibinfo{volume}{139} of
  \emph{\bibinfo{series}{Proceedings of Machine Learning Research}},
  \bibinfo{pages}{9168--9178} (\bibinfo{year}{2021}).

\bibitem{chen2011energy}
\bibinfo{author}{Chen, G.}, \bibinfo{author}{Chac{\'o}n, L.} \&
  \bibinfo{author}{Barnes, D.~C.}
\newblock \bibinfo{journal}{\bibinfo{title}{An energy-and charge-conserving,
  implicit, electrostatic particle-in-cell algorithm}}.
\newblock {\emph{\JournalTitle{Journal of Computational Physics}}}
  \textbf{\bibinfo{volume}{230}}, \bibinfo{pages}{7018--7036}
  (\bibinfo{year}{2011}).

\bibitem{chen2015multi}
\bibinfo{author}{Chen, G.} \& \bibinfo{author}{Chac{\'o}n, L.}
\newblock \bibinfo{journal}{\bibinfo{title}{A multi-dimensional, energy-and
  charge-conserving, nonlinearly implicit, electromagnetic {V}lasov--{D}arwin
  particle-in-cell algorithm}}.
\newblock {\emph{\JournalTitle{Computer Physics Communications}}}
  \textbf{\bibinfo{volume}{197}}, \bibinfo{pages}{73--87}
  (\bibinfo{year}{2015}).

\bibitem{miller2019imex}
\bibinfo{author}{Miller, S.~T.} \emph{et~al.}
\newblock \bibinfo{journal}{\bibinfo{title}{{IMEX} and exact sequence
  discretization of the multi-fluid plasma model}}.
\newblock {\emph{\JournalTitle{Journal of Computational Physics}}}
  \textbf{\bibinfo{volume}{397}}, \bibinfo{pages}{108806}
  (\bibinfo{year}{2019}).

\bibitem{Lorenz_1992}
\bibinfo{author}{Lorenz, E.~N.}
\newblock \bibinfo{journal}{\bibinfo{title}{The slow manifold --- what is it?}}
\newblock {\emph{\JournalTitle{J. Atmos. Sci.}}} \textbf{\bibinfo{volume}{49}},
  \bibinfo{pages}{2449--2451} (\bibinfo{year}{1992}).

\bibitem{Lorenz_1987}
\bibinfo{author}{Lorenz, E.~N.} \& \bibinfo{author}{Krishnamurthy, V.}
\newblock \bibinfo{journal}{\bibinfo{title}{On the nonexistence of a slow
  manifold}}.
\newblock {\emph{\JournalTitle{J. Atmos. Sci.}}} \textbf{\bibinfo{volume}{44}},
  \bibinfo{pages}{2940--2950} (\bibinfo{year}{1987}).

\bibitem{Lorenz_1986}
\bibinfo{author}{Lorenz, E.~N.}
\newblock \bibinfo{journal}{\bibinfo{title}{On the existence of a slow
  manifold}}.
\newblock {\emph{\JournalTitle{J. Atmos. Sci.}}} \textbf{\bibinfo{volume}{43}},
  \bibinfo{pages}{1547--1557} (\bibinfo{year}{1986}).

\bibitem{MacKay_2004}
\bibinfo{author}{MacKay, R.~S.}
\newblock \bibinfo{title}{Slow manifolds}.
\newblock In \emph{\bibinfo{booktitle}{Energy Localization and Transfer}},
  vol.~\bibinfo{volume}{22} of \emph{\bibinfo{series}{Advanced Series in
  Nonlinear Dynamics}}, \bibinfo{pages}{149--192} (\bibinfo{publisher}{World
  Scientific}, \bibinfo{year}{2004}).

\bibitem{Burby_Klotz_2020}
\bibinfo{author}{Burby, J.~W.} \& \bibinfo{author}{Klotz, T.~J.}
\newblock \bibinfo{journal}{\bibinfo{title}{Slow manifold reduction for plasma
  science}}.
\newblock {\emph{\JournalTitle{Comm. Nonlin. Sci. Numer. Simul.}}}
  \textbf{\bibinfo{volume}{89}}, \bibinfo{pages}{105289}
  (\bibinfo{year}{2020}).

\bibitem{McInerney2013}
\bibinfo{author}{McInerney, A.}
\newblock \emph{\bibinfo{title}{First Steps in Differential Geometry:
  {R}iemannian, Contact, Symplectic}}.
\newblock Undergraduate Texts in Mathematics (\bibinfo{publisher}{Springer New
  York}, \bibinfo{year}{2013}).

\bibitem{Lang1999}
\bibinfo{author}{Lang, S.}
\newblock \emph{\bibinfo{title}{Fundamentals of Differential Geometry}}, vol.
  \bibinfo{volume}{191} of \emph{\bibinfo{series}{Graduate Texts in
  Mathematics}} (\bibinfo{publisher}{Springer-Verlag, New York},
  \bibinfo{year}{1999}).

\bibitem{MaRa1999}
\bibinfo{author}{Marsden, J.} \& \bibinfo{author}{Ratiu, T.}
\newblock \emph{\bibinfo{title}{Introduction to mechanics and symmetry}},
  vol.~\bibinfo{volume}{17} of \emph{\bibinfo{series}{Texts in Applied
  Mathematics}} (\bibinfo{publisher}{Springer-Verlag}, \bibinfo{address}{New
  York}, \bibinfo{year}{1999}), \bibinfo{edition}{second} edn.

\bibitem{Weinstein1971}
\bibinfo{author}{Weinstein, A.}
\newblock \bibinfo{journal}{\bibinfo{title}{Symplectic manifolds and their
  {L}agrangian submanifolds}}.
\newblock {\emph{\JournalTitle{Advances in Mathematics}}}
  \textbf{\bibinfo{volume}{6}}, \bibinfo{pages}{329--346},
  \doiprefix\url{https://doi.org/10.1016/0001-8708(71)90020-X}
  (\bibinfo{year}{1971}).

\bibitem{Turaev2002}
\bibinfo{author}{Turaev, D.}
\newblock \bibinfo{journal}{\bibinfo{title}{Polynomial approximations of
  symplectic dynamics and richness of chaos in non-hyperbolic area-preserving
  maps}}.
\newblock {\emph{\JournalTitle{Nonlinearity}}} \textbf{\bibinfo{volume}{16}},
  \bibinfo{pages}{123--135}, \doiprefix\url{10.1088/0951-7715/16/1/308}
  (\bibinfo{year}{2002}).

\bibitem{AbMa1978}
\bibinfo{author}{Abraham, R.} \& \bibinfo{author}{Marsden, J.~E.}
\newblock \emph{\bibinfo{title}{Foundations of mechanics}}
  (\bibinfo{publisher}{Benjamin/Cummings Publishing Co. Inc. Advanced Book
  Program}, \bibinfo{address}{Reading, Mass.}, \bibinfo{year}{1978}).

\bibitem{Ar1989}
\bibinfo{author}{Arnol{\cprime}d, V.~I.}
\newblock \emph{\bibinfo{title}{Mathematical methods of classical mechanics}},
  vol.~\bibinfo{volume}{60} of \emph{\bibinfo{series}{Graduate Texts in
  Mathematics}} (\bibinfo{publisher}{Springer-Verlag}, \bibinfo{address}{New
  York}, \bibinfo{year}{1989}), \bibinfo{edition}{second} edn.
\newblock \bibinfo{note}{Translated from the Russian.}

\bibitem{BurbySquire2020}
\bibinfo{author}{Burby, J.~W.} \& \bibinfo{author}{Squire, J.}
\newblock \bibinfo{journal}{\bibinfo{title}{General formulas for adiabatic
  invariants in nearly periodic {H}amiltonian systems}}.
\newblock {\emph{\JournalTitle{Journal of Plasma Physics}}}
  \textbf{\bibinfo{volume}{86}}, \bibinfo{pages}{835860601},
  \doiprefix\url{10.1017/S002237782000080X} (\bibinfo{year}{2020}).

\bibitem{Teshima2020}
\bibinfo{author}{Teshima, T.} \emph{et~al.}
\newblock \bibinfo{journal}{\bibinfo{title}{Coupling-based invertible neural
  networks are universal diffeomorphism approximators}}.
\newblock {\emph{\JournalTitle{Advances in Neural Information Processing
  Systems}}} \textbf{\bibinfo{volume}{33}}, \bibinfo{pages}{3362--3373}
  (\bibinfo{year}{2020}).

\bibitem{Raymond_1968}
\bibinfo{author}{Raymond, F.}
\newblock \bibinfo{journal}{\bibinfo{title}{Classification of the actions of
  the circle on 3-manifolds}}.
\newblock {\emph{\JournalTitle{Transactions of the American Mathematical
  Society}}} \textbf{\bibinfo{volume}{131}}, \bibinfo{pages}{51--78}
  (\bibinfo{year}{1968}).

\bibitem{Duruisseaux2023NPMapCode}
\bibinfo{author}{Duruisseaux, V.}, \bibinfo{author}{Burby, J.~W.} \&
  \bibinfo{author}{Tang, Q.}
\newblock \bibinfo{title}{Code demonstration: Approximation of nearly-periodic
  symplectic maps via structure-preserving neural networks}.
\newblock \doiprefix\url{10.2172/1972078} (\bibinfo{year}{2023}).

\end{thebibliography}

\hfill

\section*{Acknowledgements}

Research presented in this article was supported by the U.S. Department of Energy (DOE), the Office of Science and the Office of Advanced Scientific Computing Research (ASCR). Specifically, we acknowledge funding support from ASCR for DOE-FOA-2493 ``Data-intensive scientific machine learning and analysis”. This research used resources of the National Energy Research Scientific Computing Center (NERSC), a U.S. Department of Energy Office of Science User Facility located at Lawrence Berkeley National Laboratory, operated under Contract No.~DE-AC02-05CH11231 using NERSC award ASCR-ERCAP0020162.  \\

\section*{Author contributions statement}

V.D. wrote the manuscript and conducted the numerical experiments. \\
J.W.B. supervised the project on the theoretical side. \\
Q.T. supervised the project on the computational side, and designed the diagrams in Figure~\ref{fig:diagram}. \\
All authors reviewed the manuscript. \\

\section*{Competing interests}
	The authors declare no competing interests.

\end{document}